\documentclass[lettersize,journal]{IEEEtran}
\usepackage{amsmath,amsfonts,amssymb}
\usepackage{array}
\usepackage[caption=false,font=footnotesize,labelfont=rm,textfont=rm]{subfig}
\usepackage{textcomp}
\usepackage{xpatch}
\usepackage{diagbox}
\usepackage{url}
\usepackage{verbatim}
\usepackage{graphicx}
\usepackage{cite}
\usepackage[table]{xcolor}
 \usepackage{threeparttable}
\usepackage{makecell}
\usepackage{hyperref}
\usepackage{amssymb}
\usepackage{multirow}
\usepackage{stmaryrd}
\usepackage{booktabs}
\usepackage{epstopdf}
\usepackage{wasysym}
\usepackage{pifont}
\usepackage[ruled,vlined]{algorithm2e}
\usepackage{algorithmic}
\usepackage{amsthm}
\usepackage{indentfirst}
\usepackage{soul}
\usepackage{wrapfig}
\usepackage{ragged2e}
\SetKwFor{SSS}{Server:}{}{endfor}
\SetKwFor{BBB}{$S_2$:}{}{endfor}
\SetKwFor{clients}{Node:}{}{endfor}

\newtheorem{myDef}{Definition}

\newtheorem{thm}{\bf Theorem}
\newtheorem{Remark}{Remark}
\usepackage{color, xcolor}
\begin{document}
\title{Towards Performance-Enhanced Model-Contrastive Federated Learning using Historical Information \\ in Heterogeneous Scenarios}

\author{Hongliang Zhang, Jiguo Yu, Guijuan Wang,  Wenshuo Ma, Tianqing He, Baobao Chai, Chunqiang Hu

\thanks{This work was partially supported by the Major Program of Shandong Provincial Natural Science Foundation for the Fundamental Research under Grant ZR2022ZD03, NSF of China under Grants 62272256, 62202250, and 62372092, the Natural Science Foundation of Sichuan Province under Grant  2025ZNSFSC0512, the Colleges and Universities 20 Terms Foundation of Jinan City under Grant 202228093, and the Shandong Province Youth Innovation Team Project under Grant 2024KJH032. ($\textit{Corresponding author}$: $\textit{Jiguo Yu}$.)}

\thanks{H. Zhang, G. Wang, and W. Ma are with the Key Laboratory of Computing Power Network and Information Security, Ministry of Education, Shandong Computer Science Center, Qilu University of Technology (Shandong Academy of Sciences), Jinan, 250353, China, Email: hongliangzhang2022@163.com, guijuan$\_$wang@126.com, weimws@foxmail.com.}

\thanks{J. Yu and T. He are with are with the School of Information and Software Engineering, University of Electronic Science and Technology of China, Chengdu, 610054, China, Email: nstlxfh@gmail.com; jiguoyu@sina.com, sunny.he@std.uestc.edu.cn.}

\thanks{B. Chai is with the School of Computer Science and Engineering, Shandong University of Science and Technology,  Jinan, 266590, China, Email:  bbchai$\_$915@sdust.edu.cn.}

\thanks{C. Hu is with the School of Big Data and Software Engineering, Chongqing University,  Chongqing, 400044, China, Email:  chu@cqu.edu.cn.}

}

\markboth{}%
{Shell \MakeLowercase{\textit{et al.}}: A Sample Article Using IEEEtran.cls for IEEE Journals}


\maketitle
\begin{abstract}
Federated Learning (FL) enables multiple nodes to collaboratively train a model without sharing raw data.
However, FL systems are usually deployed in heterogeneous scenarios, where nodes differ in both data distributions and participation frequencies, which undermines the FL performance.
To tackle the above issue, this paper proposes PMFL, a   performance-enhanced model-contrastive federated learning framework using historical training information.
Specifically, on the node side, we design a novel model-contrastive term   into the node optimization objective by incorporating  historical local models to capture stable contrastive points, thereby  improving the consistency of model updates in heterogeneous data distributions.
 On the server side, we utilize  the cumulative participation count of each node to adaptively adjust its  aggregation weight, thereby correcting the bias in the global objective caused by different node participation frequencies.
Furthermore, the updated global model incorporates historical global models to reduce its   fluctuations in  performance between  adjacent rounds.
Extensive experiments demonstrate that PMFL achieves superior   performance compared with existing FL methods in heterogeneous scenarios.
\end{abstract}


\section{INTRODUCTION}\label{12111919}

Federated Learning \cite{mcmahan2017communication} (FL)  is a distributed  paradigm that enables multiple nodes to collaboratively train a high-accuracy global model.
Since the  data originates  from  edge nodes,  it inherently exhibits Non-Independent and Identically Distributed (Non-IID) nature, thereby leading to data heterogeneity in FL \cite{10468591}\cite{11192608}.
The data heterogeneity causes inconsistent model updates across nodes, significantly  degrading the global model's performance.
Besides, existing studies observe that FL is prone to participation heterogeneity due to the distributed deployment of nodes \cite{wang2023lightweight,jhunjhunwala2022fedvarp,gu2021fast}.
Specifically, it is common in FL to select  a subset of nodes to participate  in each round of local training (i.e., partial participation) \cite{10108910}\cite{10001832}.
However,  due to higher-priority tasks   or unexpected failures, some selected nodes may drop out in certain rounds.
This leads to different actual participation frequencies across nodes, known as  participation heterogeneity \cite{10138783}\cite{10528890}.
The study  in \cite{wang2023lightweight} proves  that  participation heterogeneity biases the global model toward the local objectives of frequently participating nodes,   thereby making the aggregated global model dominated by these nodes. 
Notably, in practice, FL is inevitably deployed in heterogeneous scenarios where both data heterogeneity and participation heterogeneity coexist, posing significant challenges for designing effective FL algorithms.
However, current FL studies   focus on only one type of heterogeneity, lacking  the ability to  tackle  both data and participation heterogeneity.

\captionsetup[subfloat]{farskip=0pt}
\begin{figure}[tbp]
\subfloat[CNN]
  {\label{111816121}  \includegraphics[width=0.45\linewidth]{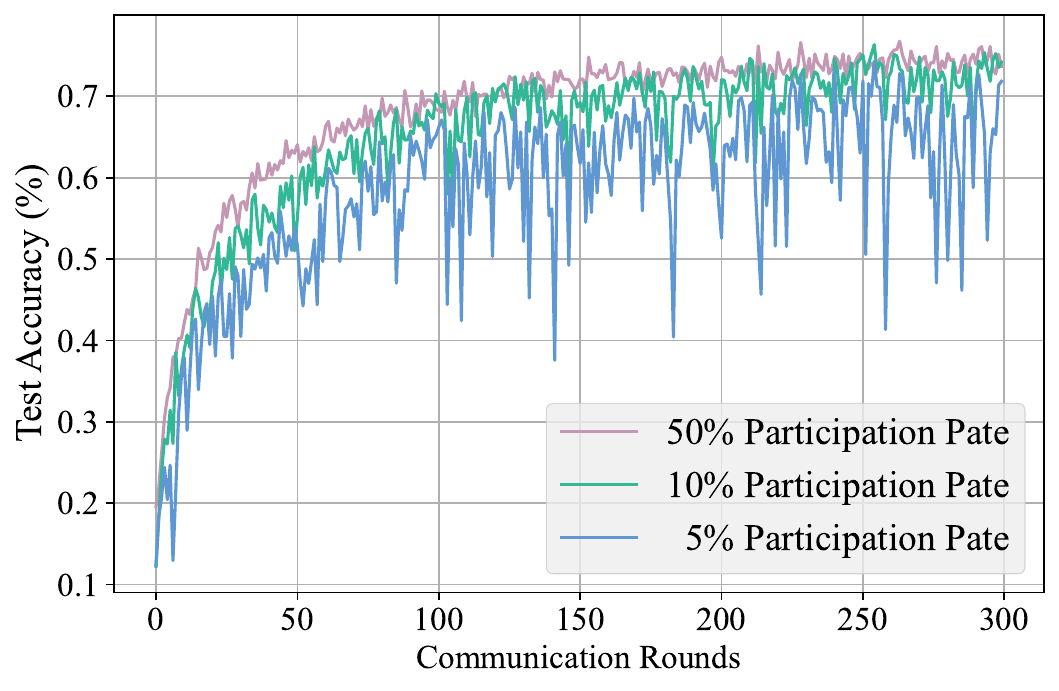}
  }
  \subfloat[ResNet18]
  {\label{111816122}  \includegraphics[width=0.45\linewidth]{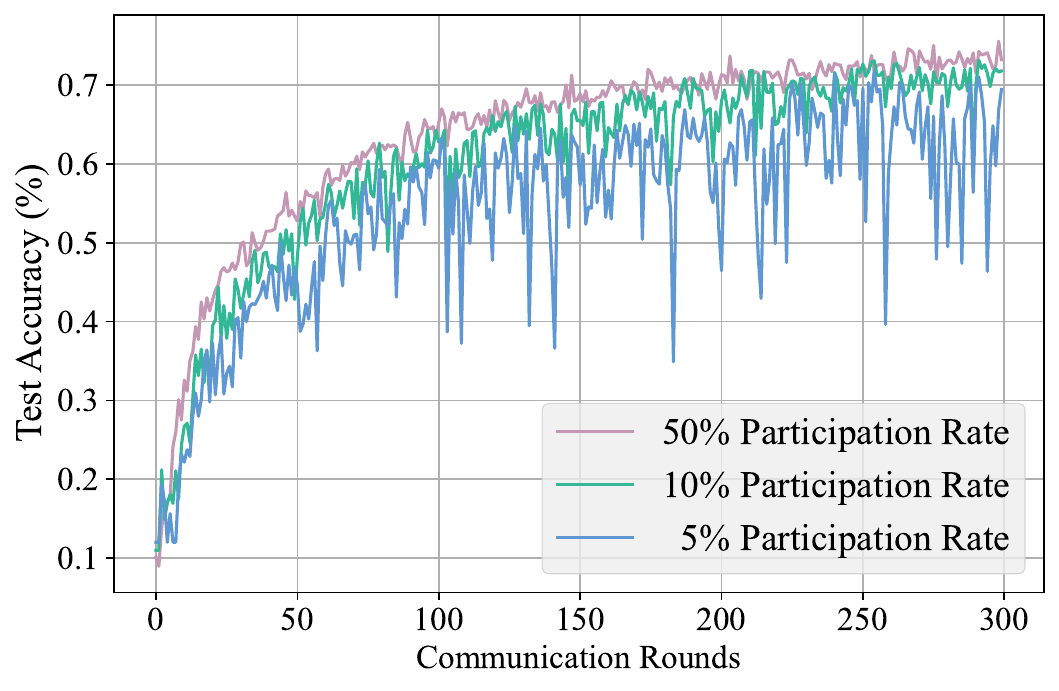}
  }
  \caption{Test accuracy of  global model in FL under varying  participation  rates on CIFAR10 \cite{krizhevsky2009learning} using average aggregation   in the Non-IID setting. Results are shown for two  model architectures: (a) CNN and (b) ResNet18.
}\label{11522361}
\end{figure}

To tackle  data heterogeneity,  numerous  works \cite{li2021model,11202428,miao2023secure,Seo_2024_CVPR,10413546,10286887} have proposed model-contrastive FL methods, which incorporate contrastive learning \cite{khosla2020supervised} into the FL optimization.
Specifically, these works \cite{li2021model,11202428,miao2023secure,Seo_2024_CVPR,10413546,10286887} maximize the similarity between the representation extracted by the current local model and the positive sample (i.e., the representation  extracted by the global model) while minimizing its similarity to the negative one (i.e., the representation  extracted by   the previous-round local model)  for the same input.
This promotes alignment between the local and global models,  improving the consistency of model updates across nodes under Non-IID data.
However, during local iterative optimization, these works \cite{li2021model,11202428,miao2023secure,Seo_2024_CVPR,10413546,10286887} compute the contrastive loss using only a single positive sample and a single negative sample as the contrastive points.
This leads to instability of  contrastive points across local training process, thereby reducing the effectiveness of contrastive learning.
Moreover, as shown in Fig. \ref{11522361}, we  observe  that   the  performance of global model  fluctuates sharply  between  adjacent rounds  under partial  participation, especially at low participation rates.
These fluctuations further exacerbate the instability of  contrastive points, making it difficult for the global model to provide stable  representations during local training, further weakening contrastive learning's ability.
Meanwhile,  participation heterogeneity not only  biases the global model toward the local objectives of frequently participating nodes, but also leads to actual participation rates being  lower than  the preset   threshold, further exacerbating fluctuations in the global model's performance.
Therefore, we raise the question: \textit{Is it possible to design an effective  model-contrastive FL that  tackles the issues caused by data and participation heterogeneity outlined above?
}

To answer this question, this paper proposes a \underline{\textbf{P}}erformance-enhanced \underline{\textbf{M}}odel-contrastive \underline{\textbf{F}}ederated \underline{\textbf{L}}earning framework, termed PMFL,  which leverages historical information to  boost  the FL performance  under  heterogeneous scenarios. 
The key novelty of PMFL lies in: $(\textbf{\text{i}})$ 
A novel model-contrastive term is designed in the optimization objective. It incorporates multiple representations from historical local models and employs cosine distance to automatically categorize them to positive and negative samples, thereby constructing  stable contrastive points.
$(\textbf{\text{ii}})$ The aggregation weight of each node is adaptively adjusted based on  its cumulative participation count, to correct the bias in the global objective caused by participation heterogeneity.
$(\textbf{\text{iii}})$ The historical global models  are incorporated into  the global update, thereby reducing fluctuations in global model performance  under partial participation.

In summary, our main contributions are four-fold:

\begin{itemize}
  \item 
      To the best of our knowledge, we  are the first to focus on the FL challenges caused by the coexistence of data heterogeneity and participation heterogeneity, and to propose a performance-enhanced model-contrastive FL framework (PMFL)   to address the challenges.
      
    \item  
        We integrate   historical local model information into  the model-contrastive term of the node's optimization function,   thereby improving FL performance in Non-IID data.
        
  \item We  propose an  aggregation strategy that adaptively adjusts aggregation weights and incorporates historical global models to alleviate the impact of participation heterogeneity at low participation rates.
  \item 
  Extensive experiment results on multiple datasets  demonstrate that   PMFL  outperforms various existing  methods in diverse heterogeneous scenarios.
\end{itemize}

The rest of the paper is organized as follows. 
In Section \ref{6231437}, we review related works about  FL with heterogeneity. Section \ref{6231438} presents the preliminary concepts of FL. 
Details of the PMFL method are presented in Section \ref{1523409}. The experimental setup  and results are shown in Section \ref{6231439} and \ref{6231440}, respectively.
Finally, Section \ref{62314441} concludes the paper.
\section{RELATED WORKS}\label{6231437}
 This section discusses the FL challenges posed by data and participation heterogeneity, and reviews FL works related to contrastive learning.
 
\subsection{Federated Learning in Data Heterogeneity}
Federated learning is a communication-efficient distributed learning paradigm, where multiple nodes  submit  their model updates without sharing  raw training data.
However,  since each node's local dataset  is not sampled from the global joint distribution, their optimization objectives diverge in direction \cite{NEURIPS2020564127c0}, which significantly degrades the  performance  of FL.
To mitigate the impact of data heterogeneity, numerous solutions can be categorized into two primary types.
The first approach focuses on  improving  the aggregation weight strategy \cite{9847055,pmlr-v202-li23s,9425020,10854512}.
For instance, the works in \cite{9425020}  \cite{10854512} dynamically allocate  aggregation weights to  nodes by minimizing the convergence rate upper bound.
Similarly, the work  in \cite{pmlr-v202-li23s} proposes   a learnable aggregation weight method  to adjust  each node's contribution.
The second approach is to   design  regularization objectives    to guide local model training in FL \cite{li2021model}\cite{miao2023secure}\cite{Seo_2024_CVPR}\cite{zhang2025swim,11075614,10556806}.
For instance, the work in \cite{miao2023secure} incorporates  the similarity between the model-level representations into the optimization objective, thereby constraining the optimization direction of local models.
Although the aforementioned  works  enhance the FL performance under data heterogeneity, their effectiveness is limited under participation heterogeneity.

\subsection{Federated Learning in  Participation Heterogeneity}

Recently, participation heterogeneity has gained  attention in FL research. 
To reduce computation  overhead, the classic FedAvg  adopts  partial participation in each round \cite{mcmahan2017communication}. 
However, the work in \cite{jhunjhunwala2022fedvarp} demonstrates that partial participation leads to performance degradation compared with full participation. 
To this end, existing works   propose  various node selection strategies to improve the performance under   partial participation   \cite{10476711,10443546,xu2025federated,10909702,10197242,10816699}.
Their effectiveness relies on the assumption that node participation follows a known  probability.
In practice, due to node independence and unpredictable failures, the server cannot  guarantee that the selected nodes will participate the local training, resulting  in heterogeneous participation patterns \cite{bonawitz2019towards}. 
Moreover, the work in \cite{wang2023lightweight}  demonstrates   that participation heterogeneity induces   bias in the global  objective.
Specifically, when certain nodes participate more frequently, the learned global model is skewed toward their local objectives,  deviating from the true global objective.

To mitigate the negative impact of participation heterogeneity,  existing studies \cite{jhunjhunwala2022fedvarp}\cite{gu2021fast} incorporate each node's most recent  model update into the current round's global update, regardless of their participation  status.
However, because many nodes may remain inactive for extended periods, integrating  their outdated updates  into the current global update may  degrade FL performance.
To this end, the work in \cite{wang2023lightweight} proposes an adaptive aggregation weight adjustment strategy based on nodes' historical participation information to address participation heterogeneity.
In addition, several FL variants have been proposed to accelerate convergence under participation heterogeneity \cite{ying2025exactlinearconvergencefederated, NEURIPS2024_bcaebb60,weng2025heterogeneityawareclientsamplingunified}.
Although these solutions have made  progress in tackling participation heterogeneity, the coexistence of data heterogeneity and participation heterogeneity continues to cause performance degradation in FL.

\subsection{Contrastive Learning}
Contrastive learning has emerged as a prominent  research area, achieving remarkable effectiveness in learning high-quality feature representations from unlabeled data \cite{liu2021self}\cite{zheng2024heterogeneous}.
Its core idea  is to minimize the   distance between representations of different  augmented views of the same sample (positive pairs), while   maximizing the distance between  representations of augmented views of different samples (negative pairs).
A notable study first incorporates contrastive learning  into federated learning and  proposes a model-contrastive FL, named MOON \cite{li2021model}.
Subsequently, the works \cite{miao2023secure,Seo_2024_CVPR,10413546} are proposed as variations of  MOON.
For instance, the work in \cite{10413546} proposes a dense contrastive-based FL that leverages multi-scale representations with dense features to learn  more expressive representations.
However, these works  \cite{11202428, miao2023secure,Seo_2024_CVPR,10413546}  use only a single positive sample from the global model and a single negative sample from the previous-round local model as contrastive points, which makes it difficult to capture stable contrastive points across local training.
In addition,  under participation heterogeneity with low participation rates, the   performance of the global model  fluctuates sharply across adjacent rounds, thereby exacerbating the instability of  contrastive points.
Thus, there is a current research gap in  addressing the above challenges under  heterogeneous   scenarios.
To bridge this gap, we propose a  performance-enhanced model-contrastive FL framework to tackle  the above issues.


\section{PRELIMINARIES} \label{6231438}
The FL  system usually   consists of a server and $K$ nodes.
Let $\mathcal{K}$ represent  the set of nodes, with the local dataset of node $k$ denoted as $\mathcal{D}_k$, and the  overall dataset  given by $\mathcal{D} := \bigcup_{k\in\mathcal{K} } \mathcal{D}_k$.
The objective of FL is to learn a   global model $W$ on the dataset $\mathcal{D}$  to solve the following optimization problem:
\begin{equation}\scalebox{1}{$
W^\star = \arg\min_{W} \frac{1}{K}\sum_{k\in\mathcal{K}} F(  \mathcal{D}_k, W)$},\label{3231151}
\end{equation}
where  $F(\mathcal{D}_k,W) := \mathbb{E}_{(x,y)\sim\mathcal{D}_k}[l((x,y),W)]$ represents  the local objective of node $k$,   $l(\cdot)$ denotes the  (local) empirical risk, and $W^\star$ is the optimal global model.
In the $t$-th  round, each selected node $k$ trains  the received global model $W^t$      on its local data to obtain the updated local model $w^t_k$. The node   then   submits its  model update  $\Delta^t_k := w^t_k-W^t$ to the server. The global model $W^{t}$ is updated as follows:
\begin{equation}\scalebox{1}{$
W^{t+1}= W^{t}+h(\Delta^t_1,\Delta^t_2,\cdots,\Delta^t_k),$}  
\end{equation}
where $h(\cdot)$ denotes the function used to  calculate  the global update.
However, data heterogeneity leads to inconsistent  model updates across nodes, which  degrades the FL performance.
In addition, participation heterogeneity causes  the global model to  learn toward  the optimization objective of  frequently participating nodes.
The study  in \cite{wang2023lightweight} formalizes the relationship between  nodes' participation frequency    and   the global  objective,  as stated   in Theorem \ref{3302334}.

\begin{thm} \label{3302334}
  Given that   the node $k$ participates in each round with probability $p_k$, and its aggregation  weight $x^t_k$ is time-constant, i.e., $x^t_k = x_k$, but generally $x_k$ may not be equal to $x_{k^\prime}$, the  FL minimizes the following    objective:
  \begin{equation}\scalebox{1}{$
    W^\star = \arg\min \frac{1}{Y}\sum_{k\in\mathcal{K}} x_k p_k F(\mathcal{D}_k,W),$}\label{2231717}
  \end{equation}
  where $Y=\sum_{k\in\mathcal{K}}x_k p_k$.
\end{thm}

Theorem \ref{3302334} indicates that  when all nodes participate with the same frequency $($i.e.$, p = p_k, \forall k)$,   FL converges to an  objective weighted by $x_k$,  which  differs from the objective defined in formula (\ref{3231151}).
However,  if all nodes are assigned the same aggregation weight $x_k$, the global model  becomes inherently biased toward nodes that participate more frequently.
Furthermore,   in practice,  node participation frequencies are typically unknown, making it difficult to define a fixed optimization objective for FL.
Thus, it is necessary to design an adaptive strategy to adjust  the aggregation weight $x_k$   for minimizing the objective in formula (\ref{3231151}) 
under  participation heterogeneity.

\section{The  Design of PMFL} \label{1523409}
This section provides a high-level overview of the  PMFL framework,  followed by a detailed description.
\subsection{High-Level Description of PMFL}
The PMFL framework consists of a central server and multiple nodes.
The server collaborates with the nodes to perform federated training.
In each round,  PMFL  performs the following three phases.
\begin{itemize}
  \item \textit{Phase I. Node Selection:}
The server randomly selects  a subset  of nodes for the current round, and  distributes the latest global model parameters to the selected nodes.
  \item \textit{Phase II. Local Model Training:}
If a node participates in training,  it initializes  its local model using the received global model parameters   and  fine-tunes the  model  by optimizing the local objective.
After completing the training,  each  participating node  sends  its model update  to server.
  \item \textit{Phase III. Aggregation:}
The server aggregates the submitted model updates to calculate the updated global model.
\end{itemize}

The process from \textit{Phase I} to \textit{Phase III} is repeated until the predetermined $T$ rounds are completed.
Notably, we establish a local sliding buffer at each node to store its historical local models, and a global sliding buffer on the server to store historical global models.
During  \textit{Local Model Training},  the model-contrastive term incorporates the historical local models stored in the local  buffer,  thereby forming a novel optimization function.
In \textit{Aggregation},  the global model is computed  using each node's cumulative participation count and the historical global models stored in the global  buffer.
The detailed procedures of \textit{Phase I} to \textit{III} are presented below.

\subsection{Node Selection Phase}
To reduce computation  overhead, the server selects a subset of nodes to participate in training in each  round.
Since each node independently decides whether to participate  based on its  availability, some selected nodes may drop out.
To formally  indicate each node's participation  status, we define an indicator $\mathbb{A}_k^t$, where $\mathbb{A}_k^t=1$ if node $k$    participates  in round $t$ and $\mathbb{A}_k^t=0$ otherwise.
Thus, only nodes with $\mathbb{A}_k^t=1$   perform the \textit{Local Model Training} phase in  $t$-th round.

\subsection{Local Model Training Phase}
\textbf{Insights.} 
As shown in Algorithm \ref{6141051},    nodes   with $\mathbb{A}_k^t = 1$  fine-tune  the global model  received from the server.
However,  data heterogeneity causes the deviations  in  optimization directions across  nodes, degrading the FL performance.
Existing FL works \cite{11202428, miao2023secure,Seo_2024_CVPR,10413546} integrate a model-contrastive term into the  optimization objective to mitigate the impact of data heterogeneity.
These methods  rely on  a single global representation as the positive sample and a single previous local representation as the negative sample.
This leads to instability of  contrastive points across local iterations.
To this end, we incorporate historical local models into the model-contrastive term to provide additional positive and negative samples, thereby capturing more stable contrastive points.
The representations  learned by these historical local models are  assigned as positive or negative samples  based on their  similarities from the representation learned by the current local model under the same input. 
In the following, we first describe the model architecture and then present the optimization function.

\begin{algorithm}[t]\label{6141051}
   \begin{small}
    \caption{$\textbf {Local Model Training}$}
    \label{961011}
    \LinesNumbered
    \KwIn {$E, \mathcal{D}_k, W^t, \eta_l.$ $\quad$$\triangleright$  $\mathcal{D}_k$ is local dataset of node $k$, $E$ denotes the number of local iterations,  $W^t$ is the global model parameters, $\eta_l$ is the local learning rate.
}
    \KwOut {$\{\Delta^{t}_k\}_{k \in \mathcal{K}}$. }
 $\{\mathbb{A}_k^t\}_{k \in \mathcal{K}} \leftarrow \{0,1\}$;\\
 \For{each node $k\in \mathcal{K}$}{
 \eIf{$\mathbb{A}_k^t=1$}{
 $w^{tE}_k \leftarrow W^t$;\\
 \For{each  iteration $e \in \{0,\cdots,E-1\}$}{
   Randomly sample $\mathcal{D}_k^{t,e} \subset \mathcal{D}_k$;\\
      $g^{tE+e}_k \leftarrow \nabla F(\mathcal{D}_k^{t,e}, w^{tE+e}_k)$;\\
      $w^{tE+e+1}_k \leftarrow w^{tE+e}_k- \eta_l g^{tE+e}_k$;\\
 }
 $\Delta^{t}_k \leftarrow w^{(t+1)E}_k-  w^{tE}_k$;\\
 }
{$\Delta^{t}_k \leftarrow 0$;}
  Send $\Delta^{t}_k$ to server;\\
 }
    \end{small}
\end{algorithm}

 

\subsubsection{Model Architecture}

Each local model, parameterized by $w$, consists of an encoder  layer, a projection layer, and a classification layer, formalized as $\operatorname{Enc}(\cdot,{}^{(1)}\!w)$, $\operatorname{Pro}(\cdot,{}^{(2)}\!w)$, and $\operatorname{Cla}(\cdot,{}^{(3)}\!w)$, respectively.
Each layer is parameterized by a distinct subset of $w$.
These  subsets satisfy $w = {}^{(1)}\!w \cup {}^{(2)}\!w\cup {}^{(3)}\!w$ and ${}^{(i)}\!w \cap {}^{(i^{\prime})}\!w = \varnothing$ for all $i \neq i^{\prime}$.
\begin{itemize}
  \item Encoder Layer. This layer  is a neural network, such as a Convolutional Neural Network \cite{gu2018recent} (CNN) or Residual Network \cite{zhang2017residual} (ResNet), designed to transform input samples into an encoder vector.
  \item Projection Layer. 
  This  layer  consists of a multi-layer perceptron  and an activation function.
      It maps the encoded vector into  a projected representation vector.
      Given a local model $w$ and sample feature  $x$, the local representation vector is computed as $z =\operatorname{Pro}(\operatorname{Enc}(x,{}^{(1)}\!w),{}^{(2)}\!w)$. For simplicity, we express the relation  as $z = \operatorname{Pro}(\operatorname{Enc}(x,w),w)$.
  \item Classification layer. The  layer takes the representation vector as input and produces class probabilities  used to calculate the classification loss.
\end{itemize}

\subsubsection{Optimization Function}
Each node  fine-tunes its local model using the designed optimization function.
For classification tasks, this  function consists of a classification term $l_{cro}$ used for classification,  and a model-contrastive term $l_{con}$ designed to align the local models with the global model.
The   term $l_{con}$ leverages additional positive and negative samples to compute the contrastive loss, as detailed below.

Specifically, each node $k$ maintains a local sliding buffer $\mathcal{S}_l$ of length  $N$ that stores its most recent local models.
Let $E$ be the number of local iterations per round with $0$$\leq$$e$$\leq$$E$$-1$.
The local model of node $k$ at the $e$-th local iteration of round $t$ is denoted by  $w^{tE+e}$.
Thus,  the  local buffer  is defined as $\mathcal{S}_l := \{w^{tE+e-N}_k, \cdots,w^{tE+e-1}_k\}$.
When $\mathcal{S}_l$ is filled, the earliest  model is  replaced by the newest one.
Given a sample feature $x$, the node computes its historical representation using each model  stored in the buffer.
Formally, the set of  historical  local representation vectors is defined as:
 $$\scalebox{0.9}{$\mathcal{Z} = \left\{ z_k^j = \operatorname{Pro}(\operatorname{Enc}(x,w_k^j),w_k^j) \,\middle|\, j \in [tE + e - N,\ tE + e - 1] \right\}.$}$$
Based on the  set $\mathcal{Z}$, the node calculates the similarity  between each historical local representation vector  $z_k^j$ and the current local representation  vector $z^{tE+e}_k$.
Given a threshold $\mu$, if the similarity  is greater than or equal to $\mu$,  the vector $z_k^j$ is regarded as a positive sample and added to the positive sample set $\mathcal{Z}^+$.
Conversely,  $z_k^j$ is allocated to the negative sample set $\mathcal{Z}^-$.
Formally, the sets $\mathcal{Z}^+$ and $\mathcal{Z}^-$ are defined as:
\begin{align} 
\mathcal{Z}^+ = \left\{ z_k^j \in \mathcal{Z} \,\middle|\, \operatorname{Sim}(z_k^{tE+e}, z_k^j) \geq \mu \right\}, \nonumber\\
\mathcal{Z}^- = \left\{ z_k^j \in \mathcal{Z} \,\middle|\, \operatorname{Sim}(z_k^{tE+e}, z_k^j) < \mu \right\}, \nonumber 
\end{align}
where $\operatorname{Sim}(a,b) = a^T b/(\Vert a \Vert \Vert b\Vert)$ represents the cosine similarity between $a$ and $b$.
The similarity value closer to 1 means  that the two vectors are  aligned in direction, whereas the value closer to -1 indicates they are  opposite. 
The threshold $\mu$ is  defined to $\operatorname{Sim}(z_k^{tE}, Z^t)$, where $Z^t$ is the global representation derived from the global model $W^t$.
This threshold  varies across nodes $k$  in each round.
The two sample sets satisfy  $\mathcal{Z}^+ \cup \mathcal{Z}^- = \mathcal{Z}$ and  $\mathcal{Z}^+ \cap \mathcal{Z}^- = \emptyset$.
Notably, at the beginning of each iteration,  $\mathcal{Z}^+$ and $\mathcal{Z}^-$ are reset to empty sets.
To construct stable contrastive points, we aim to minimize the distance between  $z_k^{tE+e}$ and the representations in ${Z^t}\cup \mathcal{Z}^+$, while maximizing its distance from the representations in $\mathcal{Z}^-$.
Thus, similar to \cite{li2021model}, the model-contrastive loss is defined as:
\begin{equation}\scalebox{1}{$
l_{con} = -\log \frac{\operatorname{pos}}{\operatorname{pos}+\operatorname{neg}},$}     
\end{equation}
where $\operatorname{pos}$ and $\operatorname{neg}$ is computed as follows:
\begin{equation}\scalebox{0.9}{$
\begin{aligned}
 \operatorname{pos} =& \operatorname{exp}\left(\frac{\operatorname{Sim}(z^{tE+e}_k,Z^t)}{\tau}\right) + \sum_{z_k^j \in \mathcal{Z}^+} \operatorname{exp}\left(\frac{\operatorname{Sim}(z^{tE+e}_k,z_k^j)}{\tau}\right),  \nonumber   
\end{aligned}$}
\end{equation}
and $\operatorname{neg} = \sum_{z_k^j \in \mathcal{Z}^-} \operatorname{exp}\left(\operatorname{Sim}(z^{tE+e}_k,z_k^j)/\tau\right)$.
Here,   $\tau$ is the temperature coefficient controlling the scaling of the similarity.
 Combining  the terms $l_{cro}$ and $l_{con}$, the loss for an input sample $(x, y)$ is given by:
\begin{equation}
\mathcal{L} = l_{cro}((x,y),w^{tE+e}_k)+\lambda l_{con}(x,W^{t},\mathcal{S}_l),  
\end{equation}
where $l_{cro}$ is the  classification loss,  
$\lambda$ is the importance  coefficient of  model-contrastive term, $x$ is the sample feature,  and $y$ is the sample label.
Formally, the local objective of  node $k$ is defined as:
\begin{equation}
\begin{aligned}
\min_{w_k} F(\mathcal{D}_k,w_k) =    \mathbb{E}_{(x,y)\sim\mathcal{D}_k} \mathcal{L}. 
\end{aligned}
\end{equation}

\subsubsection{Fine-tuned Model}
Nodes with $\mathbb{A}_k^t =1 $  fine-tune  their local models using the aforementioned model architecture and  optimization objective.
During  the $e$-th local iteration, the node $k$ randomly samples a mini-batch $\mathcal{D}_k^{t,e}$ from its local dataset $\mathcal{D}_k$, and calculates the  gradient $g_k^{tE+e}$ based on  the optimization function:
$$
g_k^{tE+e}= \nabla F(\mathcal{D}_k^{t,e},w_k^{tE+e}), 
$$
where $\nabla$ denotes the derivative operation. The local model is then updated via stochastic gradient descent:
$$
w_k^{tE+e+1}=w_k^{tE+e}-\eta_l g_k^{tE+e},
$$
where  $\eta_l$ is the local learning rate.
After completing  $E$ local iterations, the node  calculates  its model update as follows:
$$
\Delta_k^{t}=w^{(t+1)E}_k-  w^{tE}_k,
$$
and  submits $\Delta_k^{t}$ to the server.
Notably, for nodes with $\mathbb{A}_k^t = 0$, the model update $\Delta_k^{t}$ is set to zero vector.


\subsection{Aggregation Phase}

We  propose an aggregation strategy consisting of the \textit{Adaptive Aggregation Weight Calculation} step and the \textit{Global Model Aggregation} step.
First, the server  adjusts each node's aggregation weight based on its  cumulative participation count to mitigate the bias  caused by participation heterogeneity.
Second,  the server incorporates historical global models   into the global update to reduce performance fluctuations.

\subsubsection{Adaptive Aggregation Weight Calculation}
\textbf{Insights.} 
The global optimization objective defined in formula (\ref{2231717}) is equivalent to that in formula (\ref{3231151}) when   the aggregation weight $x_k$  is set to $1/p_k$ for each node $k$ under  participation heterogeneity.
Given  $T$  rounds, the relation $x_k = 1/p_k$ can be rewritten as:
\begin{equation}\scalebox{1}{$
x_k = 1/p_k = 1/\frac{M_k}{T} = T/M_k,$}\label{2241642}
\end{equation}
where $M_k$ denotes the cumulative participation count of node  $k$  over $T$ rounds.
Formula (\ref{2241642}) shows that the weight $x_k$  equals $T /M_k$, which is interpreted as the average participation  interval of node $k$ when the first interval starts at round $t$=0.
However, since  a node's future participation status  is unknown,  $M_k$ is not available a priori.
Thus,  in the $t$-th round, the  weight $x_k$ can be estimated using the current cumulative participation count of node $k$.
Nevertheless, this estimation method faces two  issues.
(\textit{i}) If a node has not   participated in any previous rounds, its average interval cannot be estimated.
(\textit{ii}) When the current  round $t$ is small and node $k$ has participated frequently, the estimated average  interval may  deviate from the ideal optimal value $T/M_k$.
To this end, we introduce  a positive integer $C$$\in$$\mathbb{N}^+$ as the ``cutoff'' interval.
Specifically, if node $k$ has not participated in any of the most recent $C$ rounds, its cumulative participation count is incremented by 1.
Using the cutoff  interval $C$ to calculate  the average participation  interval  effectively mitigates the above issues.
Notably, when  $C$ is  too large,  it is difficult to estimate the average  interval if a node has not participated for many rounds.
In contrast, a smaller $C$ may overestimate the node's actual participation frequency.
Thus, selecting an appropriate $C$ value can effectively improve the FL performance, as validated by our experimental results.


Based on the above insights, the server  calculates each node's aggregation weight using its  cumulative participation count,  as shown in Algorithm \ref{6141052}.
Let $R_k$ denote  the current  participation count (including  cutoffs) of node $k$, and let $Q_k$ denote  its current participation interval, i.e., the number of rounds elapsed since its last participation.
At the beginning of the FL  task, both $Q_k$ and $R_k$ are initialized to 0.
Specifically, in the aggregation phase, the server increments the current participation interval $Q_k$ by 1 for each node.
If  node $k$ satisfies $\mathbb{A}_k^t$=1 or $Q_k$=$C$,   its aggregation weight is updated based on its current average participation interval.
In contrast, for other nodes, the aggregation weight remains unchanged.
For nodes whose weights need to be updated, their current participation interval $Q_k$ is recorded as $Q_k^\star$.
If such a node  has never participated  in the previous $(t-1)$ rounds  (including cutoff), its aggregation weight  is set as $x_k^t\leftarrow Q^\star_k$.
Conversely, the aggregation weight of  node $k$ is updated  as:
\begin{equation}
\scalebox{1}{$x_k^t= (R_k\cdot x_k^{t-1}+Q^\star_k)/(R_k+1),$}
\end{equation}
where $x_k^t$ is interpreted as the average participation interval of node $k$ up to round $t$.
After the weight $x_k^t$ is updated,  the current participation count $R_k$ is incremented by 1 only for nodes satisfying $\mathbb{A}_k^t$=1 or $Q_k$=$C$, and the current interval  $Q_k$ is reset to 0.
Notably,  the server can obtain each node's aggregation weight   regardless of whether the node participates in the current round.


\begin{algorithm}[t]\label{6141052}
   \begin{small}
    \caption{$\textbf {Aggregation}$}
    \label{961011}
    \LinesNumbered
    \KwIn {$C, \eta_g, H, \mathbb{A}_k^t, \Delta^{t}_k, Q_k, R_k.$ $\quad$ $\triangleright$ $C$ is cutoff  interval, $\eta_g$ is the global learning rate, $H$ is the size of global sliding buffer,
 $\Delta^{t}_k$ is the   model update of node $k$, $Q_k$ is the current participation interval of node $k$,  and $R_k$ is the current participation count of node $k$.
}
    \KwOut {$W^{t+1}$.}
    // $\mathtt{Adaptive\ Aggregation\ Weight\ Calculation}$\\
\For{each   node $k$ $\in$ $ \mathcal{K}$}{
    $Q_k \leftarrow Q_k+1$;\\
\eIf{$\mathbb{A}_k^t=1$ or $Q_k= C$}{
$Q^\star_k \leftarrow Q_k$;\\
\eIf{$R_k = 0$}{$x_k^t\leftarrow Q^\star_k$;\\}{$x_k^t\leftarrow (R_k\cdot x_k^{t-1}+Q^\star_k)/(R_k+1)$;\\
    }
$R_k \leftarrow R_k+1$; $Q_k \leftarrow 0$;\\
}{$x_k^{t}\leftarrow x_k^{t-1}$;\\}}
    // $\mathtt{Global \ Model \ Aggregation}$\\
$W^{t+1} = (1-\psi)\left(W^t -\eta_g \sum_{k \in \mathcal{K}}x_k^t \Delta_k^t\right)+ \frac{\psi}{H-1}\sum_{h=1}^{H-1}W^{t-h}$;\\
Distribute the global model $W^{t+1}$ to nodes ;\\
    \end{small}
\end{algorithm}

\subsubsection{Global Model Aggregation}
To compute the updated  global model $W^{t+1}$, the server  uses the aggregation weight    $x_k^t$ to determine each node's contribution.
Formally, the global model is updated as follows:
\begin{equation}\scalebox{1}{$
W^{t+1} = W^t -\eta_g \sum_{k \in \mathcal{K}}x_k^t \Delta_k^t,$} \label{1209130}
\end{equation}
where $\eta_g$ is the global learning rate.
Notably,   nodes that do not participate in the current round have $\Delta^t_k$=0.
Thus, they do not influence the aggregation results.
Although adjusting the aggregation weights  mitigates the global model bias caused by participation heterogeneity, the global model performance may fluctuate  when the participation rate   is low, as shown in Fig. \ref{11522361} of Section \ref{12111919}.
This fluctuation weakens the global representation extracted by the global model in  the model-contrastive term.

To alleviate performance fluctuations of the global model,  we incorporate  historical global models into the updated global  model.
Specifically, the server maintains a global sliding buffer $\mathcal{S}_g$ that stores the most recent $H$ global models, thereby preserving historical information from previous rounds.
Formally, the update rule of the global model is rewritten as:
\begin{equation}\scalebox{0.85}{$
W^{t+1} = (1-\psi)\left(W^t -\eta_g \sum_{k \in \mathcal{K}}x_k^t \Delta_k^t\right)+ \frac{\psi}{H-1}\sum_{h=1}^{H-1}W^{t-h},$} 
\end{equation}
where $\psi$ is a dynamic coefficient defined by:
\begin{equation}\scalebox{1}{$
\psi = 1/2 -t/(2(T-1)).$} 
\end{equation}
As the number of rounds increases, the global model gradually converges.
Thus, the dynamic coefficient $\psi$ can progressively reduce the reliance of the updated global model $W^{t+1}$ on historical global models.
By this step, the server obtains the   updated global model  $W^{t+1}$  for the current round.

\section{EXPERIMENTAL SETUP}\label{6231439}
This section presents the experimental datasets and model structures,  the settings for heterogeneity, and provides a detailed  hyper-parameter settings and compared  methods.
\subsection{Datasets and Model Structures}
In our  evaluation, we validate PMFL on multiple datasets, including SVHN, CIFAR10, CINIC, and CIFAR100.
All nodes  adopt the same  model architecture within each dataset.
\begin{itemize}
  \item For  SVHN \cite{netzer2011reading} dataset, the model consists of two convolutional layers used as the encoder  and  two fully connected layers used as the classifier.
  \item For  CIFAR10 \cite{krizhevsky2009learning} dataset, the model includes two convolutional layers used as the encoder  and three fully connected layers as the classifier.
      \item  For  CIFAR100 \cite{krizhevsky2009learning} and CINIC \cite{darlow2018cinic} datasets, the model consists of two convolutional layers used as the encoder and three fully connected layers as the classifier.
\end{itemize}
Each convolutional layer employs a kernel size of $3 \times 3$, followed by a Rectified Linear Unit (ReLU) function and a max-pooling layer with a kernel size of $2 \times 2$.

\subsection{Heterogeneity  Settings}\label{6161624}

To simulate data heterogeneity, we follow the existing FL works \cite{mu2024feddmc}\cite{10549523}   and employ  a Dirichlet distribution \cite{lin2016dirichlet} with concentration parameter $\alpha$ to generate the class distribution for each node, thereby constructing the Non-IID setting.
The parameter  $\alpha$ controls  the degree of data heterogeneity, with a smaller $\alpha$ indicating the higher degree of data heterogeneity.
In addition, to simulate participation heterogeneity, we generate each node's actual participation frequency  using a Dirichlet distribution with  parameter $\beta$.
A smaller $\beta$ indicates the larger   difference in participation frequencies among  nodes.

\begin{figure}[t]
  \centering \includegraphics[width=0.9\linewidth]{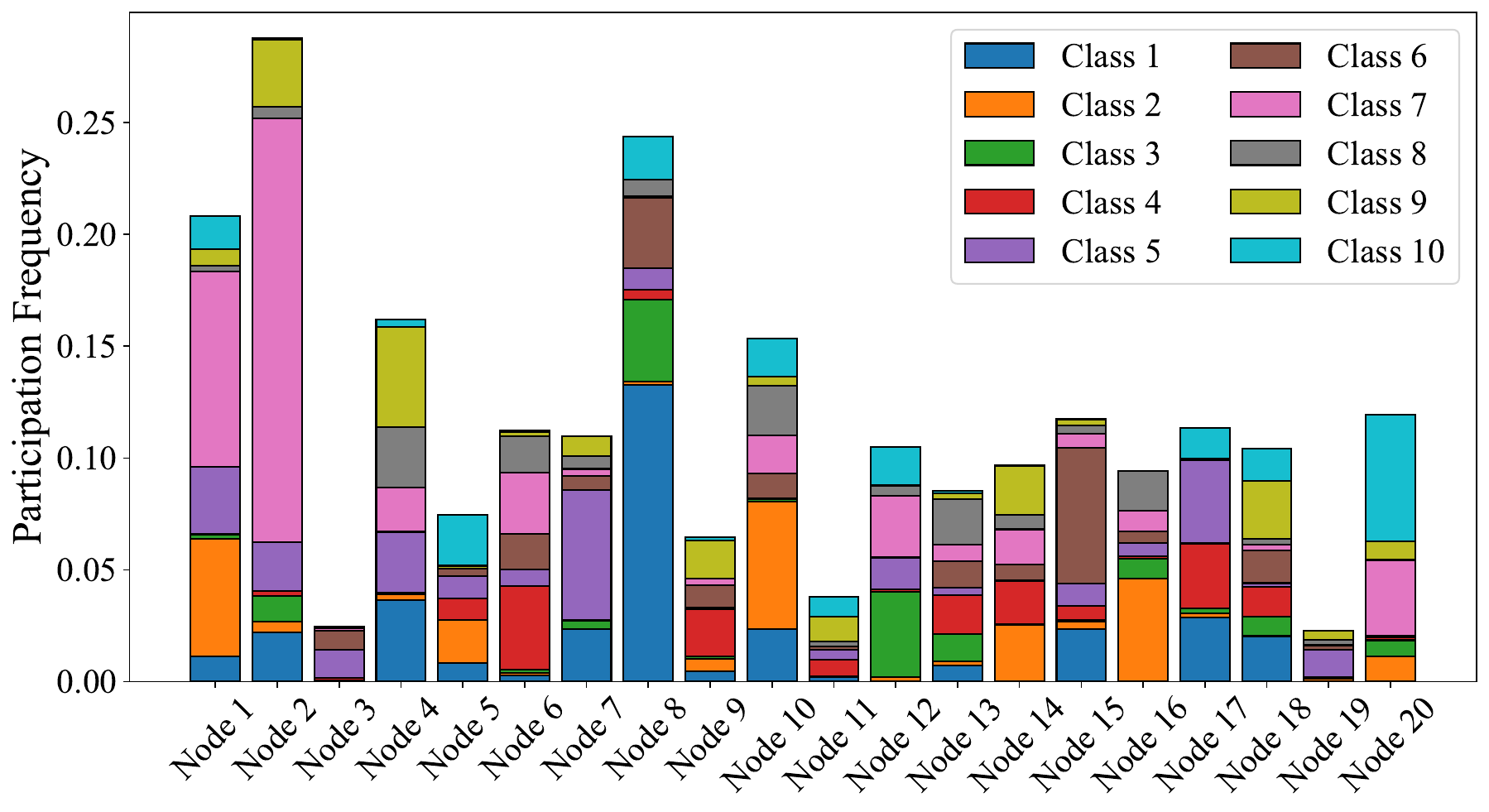}
\caption{Visualization of class distributions and participation frequencies across 20 nodes.
}
\label{1152236123}
\end{figure}

More specifically, to generate each node's participation probability, a vector $\textbf{Z}$ is sampled  from the Dirichlet distribution with parameter $\beta$, where the dimension of $\textbf{Z}$  is the same as that of each node's class distribution vector $\textbf{D}_k$.
The participation frequency of node $k$ is then calculated as:
\begin{equation}
p_k =
\begin{cases}
\frac{\langle \mathbf{Z}, \mathbf{D}_k \rangle}{r}, & \text{if } \frac{\langle \mathbf{Z}, \mathbf{D}_k \rangle}{r} \geq 0.02, \\
0.02, & \text{otherwise}. \label{6151843}
\end{cases}
\end{equation}
where $\langle \cdot,\cdot\rangle$ denotes  the inner product, and $r$ is  the normalization factor that ensures  $\mathbb{E}[p_k]= 0.1$, thereby simulating a  low participation rate in FL.
The minimum participation frequency is capped at 0.02.
For example, in a 10-class classification task,  the vectors $\textbf{D}_k$ and $\textbf{Z}$ are sampled independently  from   Dirichlet distributions with parameters $\alpha$=0.1 and $\beta$=0.1, respectively.
Assuming the  vector $\textbf{Z}$ is given by:
$$\textbf{Z} = [0.38,0.10,0.01,0.00,0.00,0.01, 0.44,0.04,0.03,0.00],$$
the participation frequency $p_k$ is  calculated using formula (\ref{6151843}).
Since  $\textbf{Z}[7] =0.44$ is the maximum value,  nodes with a higher proportion of class-7 samples tend to have higher participation frequencies.
Fig. \ref{1152236123} visualizes the relationship between data heterogeneity and participation heterogeneity.

\begin{figure}[t]
\centering
   \includegraphics[width=0.9\linewidth]{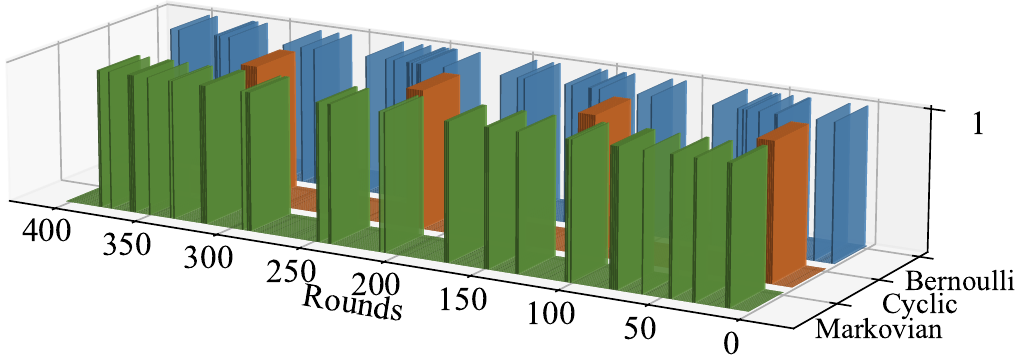}
\caption{
Visualization of various participation patterns $\mathbb{E}[p_k]= 0.1$ over the first 400 Rounds for a single node.
}
\label{101520}
\end{figure}

Although the Dirichlet distribution determines  each node's participation frequency, it does not capture the temporal dynamics of participation.
To address this, we   introduce  distinct  participation patterns in our experiments, namely bernoulli, markovian, and cyclic, to simulate   unknown participation states of each node.
Fig. \ref{101520} visualizes the three   participation patterns.
The details of these patterns are specified as follows.

\begin{itemize}
  \item  In  bernoulli   pattern, each  node  determines its participation status at each  round based on a  bernoulli distribution parameterized by $p_k$. Formally, the participation status $\mathbb{A}_k^t\in\{0,1\}$  is given by:
\begin{equation}\scalebox{1}{$
P(\mathbb{A}_k^t =1) = p_k, \quad P(\mathbb{A}_k^t =0) = 1-p_k,$} \nonumber
\end{equation}
where $P(\mathbb{A}_k^t =1)$ denotes the probability that the node participates in each round.
Since the participation status across rounds is  independent, we have $P(\mathbb{A}_k^t |\mathbb{A}_k^{t-1}) = P(\mathbb{A}_k^t)$,   ensuring   that the participation status exhibits no temporal dependency.

  \item In markovian  pattern, each node switches between participation  and  non-participation based on a two-state Markov chain. The transition probabilities are defined by:
      \begin{itemize}
      \item $p^{0\rightarrow1}_k$: Transition probability from non-participation to participation,  set by default to $0.05$. 
      \item $p^{1\rightarrow0}_k$: Transition probability from participation to  non-participation,  derived as:
          \end{itemize}
\begin{equation}
p^{1 \to 0}_k =  (1 - p_k) \cdot p^{0\rightarrow1}_k.\nonumber
\end{equation}
The transition probabilities ensure that the node's participation frequency remains stable in subsequent rounds.

\item 
In cyclic  pattern, each node alternates between participation   and non-participation within a fixed-length cycle of $T_{cycle}$ rounds.
 Specifically,  node $k$ participates in training for $p_k \times T_{cycle}$ rounds per cycle, and does not participate for the remaining $(1-p_k)\times T_{cycle}$ rounds.
To simulate the diversity of node participation, each node   begins  its cycle with a random offset $o_k\in [0,T_{cycle})$. The participation status $\mathbb{A}_k^t$ at round $t$ is formally  defined as:
\begin{equation}
    \mathbb{A}_k^t =
\begin{cases}
1, & \text{if } (t - o_k) \bmod T_{cycle} < p_k \times T_{cycle}, \\
0, & \text{otherwise}. \nonumber
\end{cases}
\end{equation}
We set the cycle length to 100 rounds.
This pattern combines  periodic and random elements, capturing the participation dynamic in real-world  scenarios.
\end{itemize}

\subsection{Hyper-Parameter Setting}
The experiments  simulate   250 nodes, each   configured with the identical  hyper-parameters.
Unless otherwise specified, the following default settings are adopted.
The local learning rate $\eta_l$ is set to 0.1, the global learning rate $\eta_g$ to 1, and the temperature parameter $\tau$ to $0.5$.
The number of local iterations $E$  is 5.
The  importance coefficient of model-contrastive term $\lambda$ is 0.5.
The global buffer size $H$ is 3, and local buffer size $N$ is 5.
The ``cutoff'' interval length $C$ is set to 50.
To simulate data heterogeneity and participation heterogeneity,  the Dirichlet parameters $\alpha$ and $\beta$ are both set to 0.1.
To simulate a scenario with low participation rate,  the expected participation frequency is set to $\mathbb{E}[p_k]$=0.1, which means that on average only 25 nodes participate in each round.
The number of  rounds $T$ is set to 1000 for  SVHN, 10000 for  CIFAR10, CINIC,  and  CIFAR100.

\subsection{Compared  Methods}
We compare our PMFL  with several baseline methods.
FedVarp \cite{jhunjhunwala2022fedvarp}  and MIFA \cite{gu2021fast}: Both methods maintain and reuse the most recent update from each node, and aggregate these cached updates during the aggregation phase to approximate the current global update.
FedHyper \cite{wang2023fedhyper}: The method is a hyper-gradient–based learning-rate scheduling algorithm for FL that adopts average aggregation.
FedAU \cite{wang2023lightweight}: This method assigns aggregation weights to model updates during aggregation based on each node's average participation interval.
FedPPO \cite{10909702}: The method evaluates each node's contribution by combining the accuracies of the global and local models, and  selects high-contribution nodes for the next round.

\section{EXPERIMENTAL RESULTS}\label{6231440}
This section reports and analyzes the performance of PMFL in heterogeneous scenarios.
\subsubsection{Performance of PMFL and Compared Methods}
\begin{table*}[t]
\centering
\caption{Accuracy results (\%)  in various methods. The best and second-best results are shown in bold and underlined, respectively.}
  \renewcommand{\arraystretch}{1.1}
\scalebox{0.8}{
\tabcolsep = 0.3cm
\begin{tabular}{c|c|c|c|c|c|c|c|c|c}
\hline\hline
\multirow{3}{*}{\makecell{ \textbf{Participation}\\  \textbf{pattern}}} & \multirow{3}{*}{\textbf{Methods}} & \multicolumn{2}{c|}{\textbf{SVHN}} & \multicolumn{2}{c|}{\textbf{CIFAR10}} & \multicolumn{2}{c|}{\textbf{CINIC}} & \multicolumn{2}{c}{\textbf{CIFAR100}} \\
\cmidrule{3-10}
 & & Training & Test & Training& Test &Training &Test &Training & Test \\
\midrule
\multirow{6}{*}{Bernoulli} & PMFL (ours) & \textbf{88.03} (+1.00) & \textbf{87.47}  (+0.58) & \textbf{86.99} (+0.97) & \textbf{78.48} (+0.50)& \textbf{68.49} (+1.15)& \textbf{64.40} (+0.98)& \textbf{54.16} (+1.43)& \textbf{45.62} (+0.58)\\
 & FedVarp \cite{jhunjhunwala2022fedvarp}& 85.73 & 85.45 & 83.06 & 77.83& 60.47 & 58.77 & 49.73 & 42.82 \\
 & MIFA \cite{gu2021fast}& 86.45 & 86.10 & 81.72 & 77.43 & 60.77 & 59.13 & 49.91 & 42.86 \\
  & FedHyper \cite{wang2023fedhyper} & 84.90 & 83.95 & 81.54 & 73.03 & 59.16 & 55.54 & 48.65 & 42.47 \\
 & FedAU \cite{wang2023lightweight}& \underline{87.03} & \underline{86.89} & \underline{86.02} & \underline{77.98}& \underline{67.34} & \underline{63.42} & \underline{52.73} & \underline{45.04} \\
 & FedPPO  \cite{10909702} & 86.95 & 86.61 & 85.85 & 77.92 & 65.64 & 62.35 & 50.80 & 43.63 \\
\midrule
\multirow{6}{*}{Markovian} &  PMFL (ours) & \textbf{88.56} (+0.92)& \textbf{87.78} (+0.77)& \textbf{86.79} (+0.82)& \textbf{78.26} (+0.42)& \textbf{67.99} (+1.12)& \textbf{64.07} (+1.02)& \textbf{52.18} (+1.51)& \textbf{44.63} (+0.67)\\
 & FedVarp \cite{jhunjhunwala2022fedvarp}& 86.73 & 86.25 & 82.52 & 77.24 & 61.09 & 59.26 & 49.45 & 42.64 \\
 & MIFA \cite{gu2021fast}& 86.86 & 86.48 & 80.86 & 76.98 & 59.71 & 57.92 & 49.83 & 42.70 \\
  & FedHyper \cite{wang2023fedhyper}& 84.34 & 82.96 & 82.01 & 73.20 &  59.74 & 55.20& 49.02 & 42.81 \\
 & FedAU \cite{wang2023lightweight} & \underline{87.64}& \underline{87.01} & \underline{85.97} & \underline{77.84}& \underline{66.87} & \underline{63.05}& \underline{50.67}& \underline{43.96} \\
 & FedPPO   \cite{10909702}& 86.25 & 85.84& 84.47& 77.40 & 61.25 & 58.79& 49.86 & 42.63 \\
\midrule
\multirow{6}{*}{Cyclic} &PMFL (ours) & \textbf{87.90} (+0.75)& \textbf{87.55} (+0.50)& \textbf{85.83} (+1.18)& \textbf{77.90} (+0.87)& \textbf{65.46} (+0.73)& \textbf{61.89} (+0.48)& \textbf{51.64} (+1.44)& \textbf{43.99} (+0.43)\\
 & FedVarp \cite{jhunjhunwala2022fedvarp}& 81.77 & 81.34 & 79.03& 72.15& 55.41 & 54.22& 46.46 & 42.13 \\
 & MIFA \cite{gu2021fast}& 76.13& 75.65& 69.50 & 67.57 & 60.77 & 59.12 & 45.08& 41.86\\
  & FedHyper \cite{wang2023fedhyper}& 84.89& 83.81 &80.80 & 72.62& 59.16 & 55.54 & 48.87 & 42.70\\
  & FedAU \cite{wang2023lightweight}& \underline{87.15} & \underline{87.05}& \underline{84.65} & \underline{77.03} & \underline{64.73}& \underline{61.41}& \underline{50.20} & \underline{43.56} \\
 & FedPPO  \cite{10909702} & 86.96 & 86.72 & 84.55 & 76.81 & 64.47 & 61.24 & 49.94 & 42.87 \\
\hline\hline
\end{tabular}}\label{323118}
\end{table*}

 The accuracy  of PMFL and the baseline methods is evaluated on SVHN, CIFAR10, CINIC, and CIFAR100.
All hyper-parameters are set to their default values.
The  results are reported  in Table \ref{323118}. 
For each column, the best-performing result in each sub-table is highlighted in bold.
Each result in the table is obtained by calculating the average   of the top-5  accuracy values from the corresponding experimental runs.
It can be observed that PMFL achieves higher   accuracy than  other methods under different participation patterns.
This superior performance is attributed to the following  three key factors.
First, PMFL incorporates historical global models into the computation of the current global model during   aggregation, thereby mitigating the performance fluctuations caused by low participation rates.
Second, it dynamically adjusts the aggregation weights based on each node's current participation count, enabling the global optimization objective to better approximate formula (\ref{3231151}).
Third, PMFL incorporates the model-contrastive  term with historical information into the local optimization function, thereby enhancing the model's ability to adapt to data heterogeneity.
Hence, these designs  enable PMFL to achieve stable and high-accuracy performance under  heterogeneous scenarios.

Notably, FedVarp and MIFA  exhibit relatively weaker performance, particularly under the cyclic pattern.
This is because they use the most recently submitted local updates from all nodes in each round.
However, when node participation frequency is low, many nodes may fail to participate in training for multiple consecutive rounds. 
As a result, the cached updates for those nodes become stale, which slows down the convergence of the global model.
Furthermore, FedHyper aggregates only the updates from currently participating nodes  and employs a simple averaging strategy. 
 Although FedHyper can adaptively adjust the local and global learning rates, its performance is significantly affected by participation heterogeneity.
In contrast, FedAU dynamically adjusts aggregation weights based on each node's cumulative participation count to correct the bias of the global optimization  objective. 
Thus, it achieves stable performance across three participation patterns.
However, since it does not address data heterogeneity, its overall performance remains inferior to that of our PMFL.
Furthermore,  FedPPO selects a subset of nodes to participate in the next  round based on their estimated contributions.
On SVHN and CIFAR10, its performance remains relatively stable across different participation patterns because  these datasets are relatively simple and  can converge  even with few participating nodes. 
However, on CINIC and CIFAR100, FedPPO's performance significantly deteriorates, indicating that contribution-based selection method cannot fundamentally resolve participation heterogeneity.

\subsubsection{Convergence of Different Methods}
\begin{figure}[!t]
\centering
\subfloat[SVHN, Bernoulli]
  {
      \label{32410582}  \includegraphics[width=0.45\linewidth]{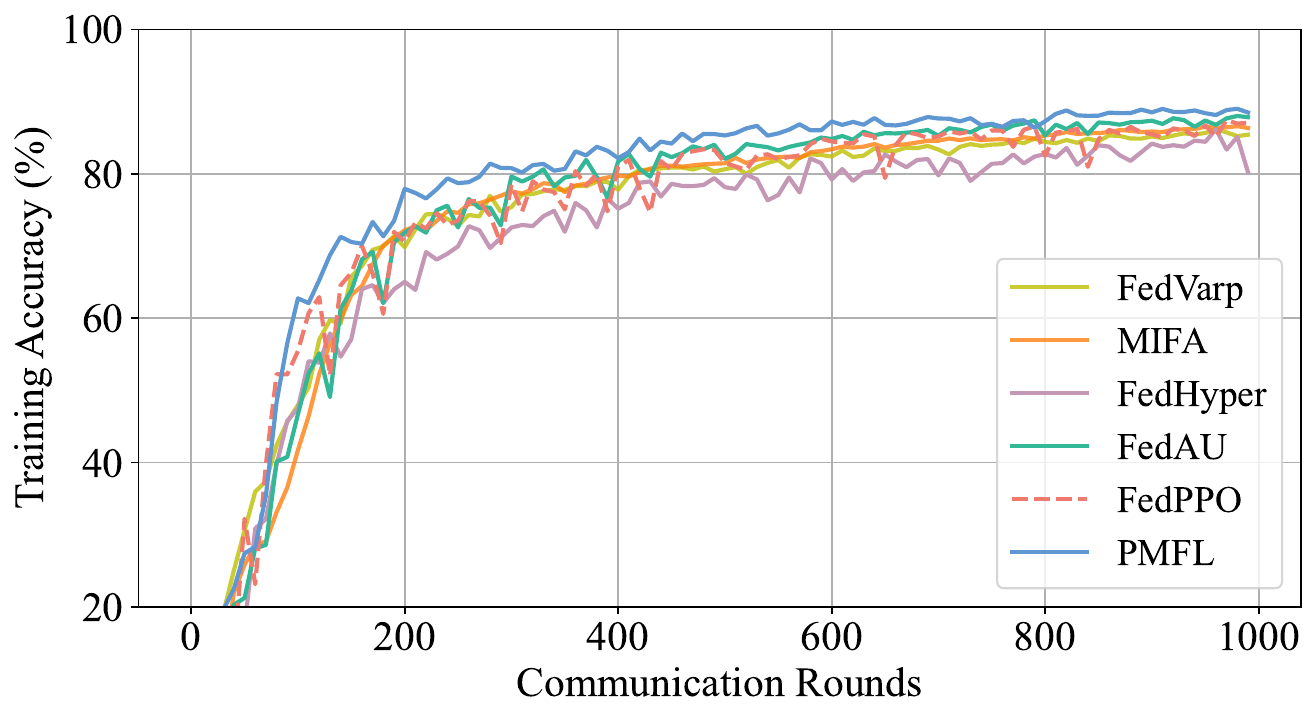}
  }
\subfloat[SVHN, Markovian]
  {
      \label{32410585}  \includegraphics[width=0.45\linewidth]{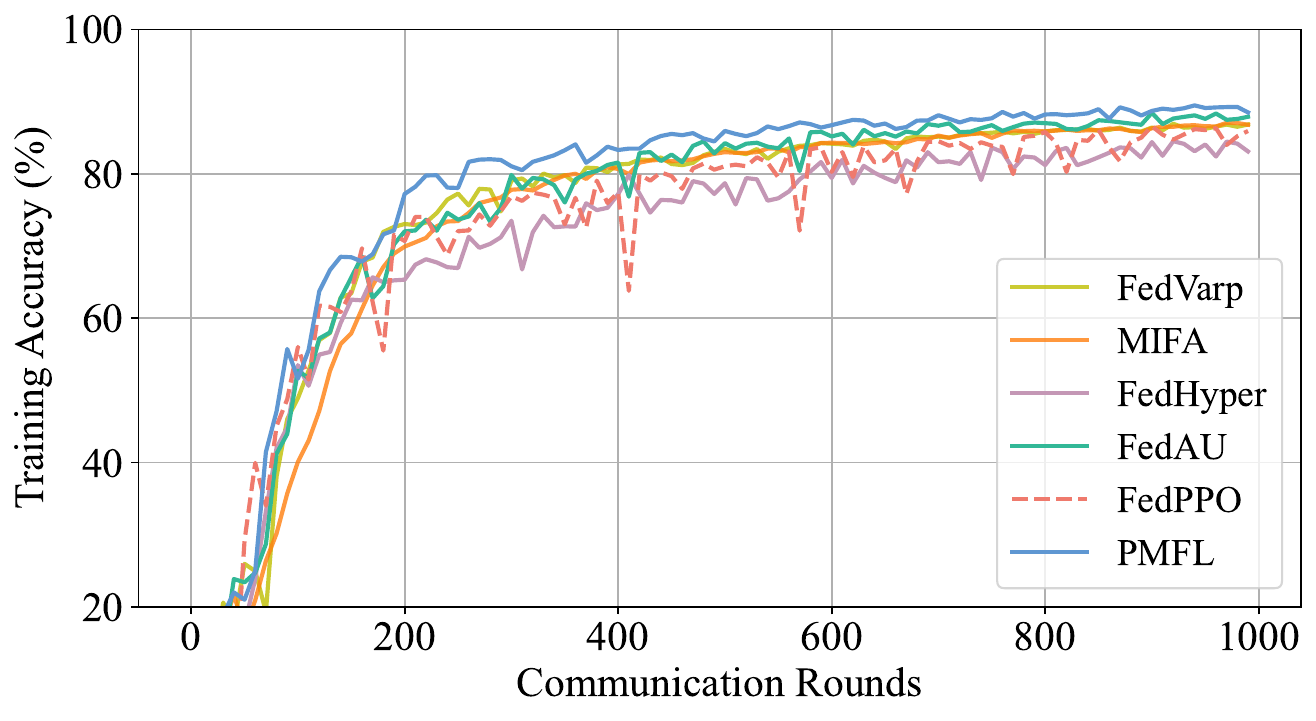}
  }

\subfloat[CIFAR10, Bernoulli]
  {
      \label{324105821}  \includegraphics[width=0.45\linewidth]{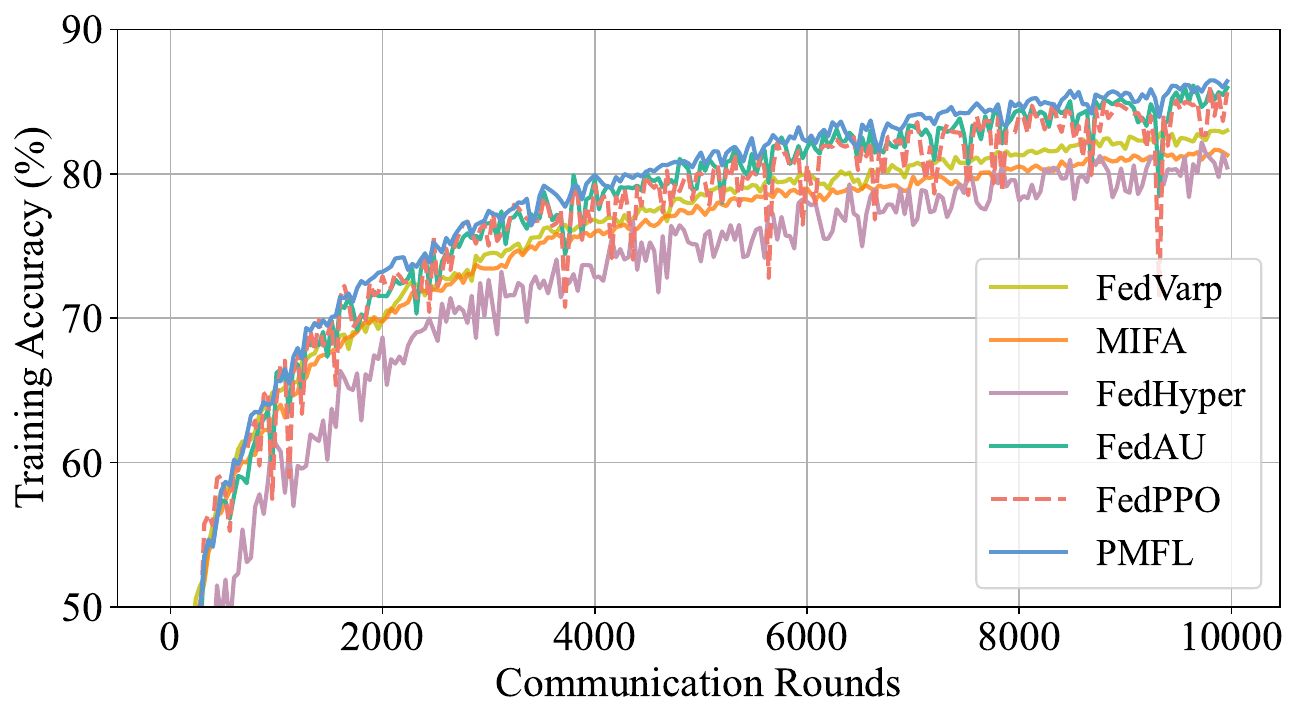}
  }
\subfloat[CIFAR10, Markovian]
  {
      \label{324105851}  \includegraphics[width=0.45\linewidth]{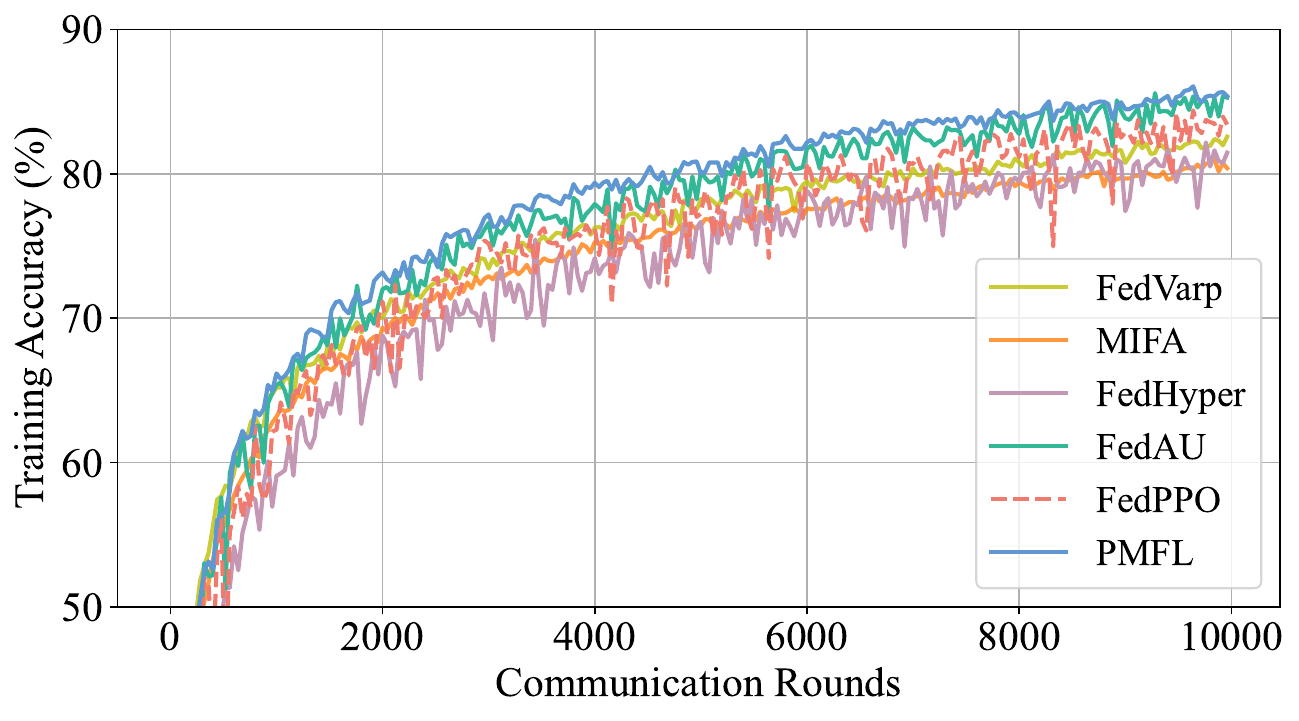}
  }

\subfloat[CINIC, Bernoulli]
  {
      \label{3241058211}  \includegraphics[width=0.45\linewidth]{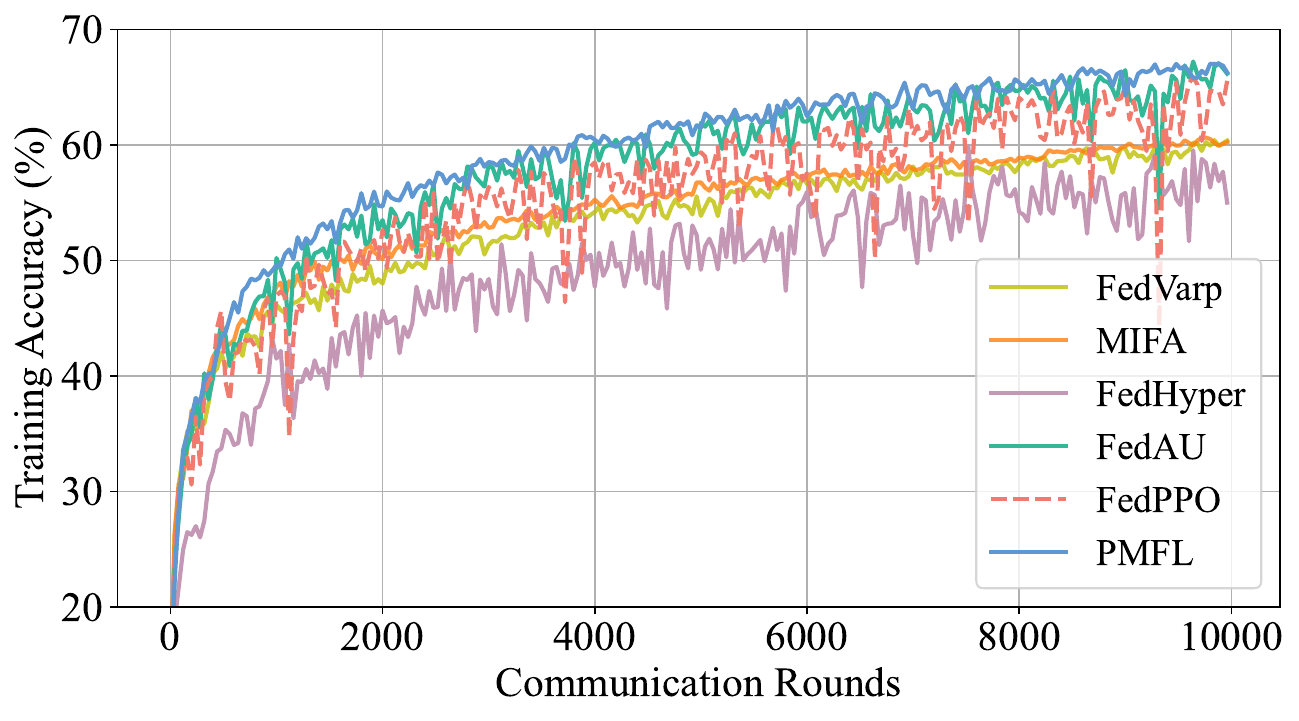}
  }
\subfloat[CINIC, Markovian]
  {
      \label{3241058511}  \includegraphics[width=0.45\linewidth]{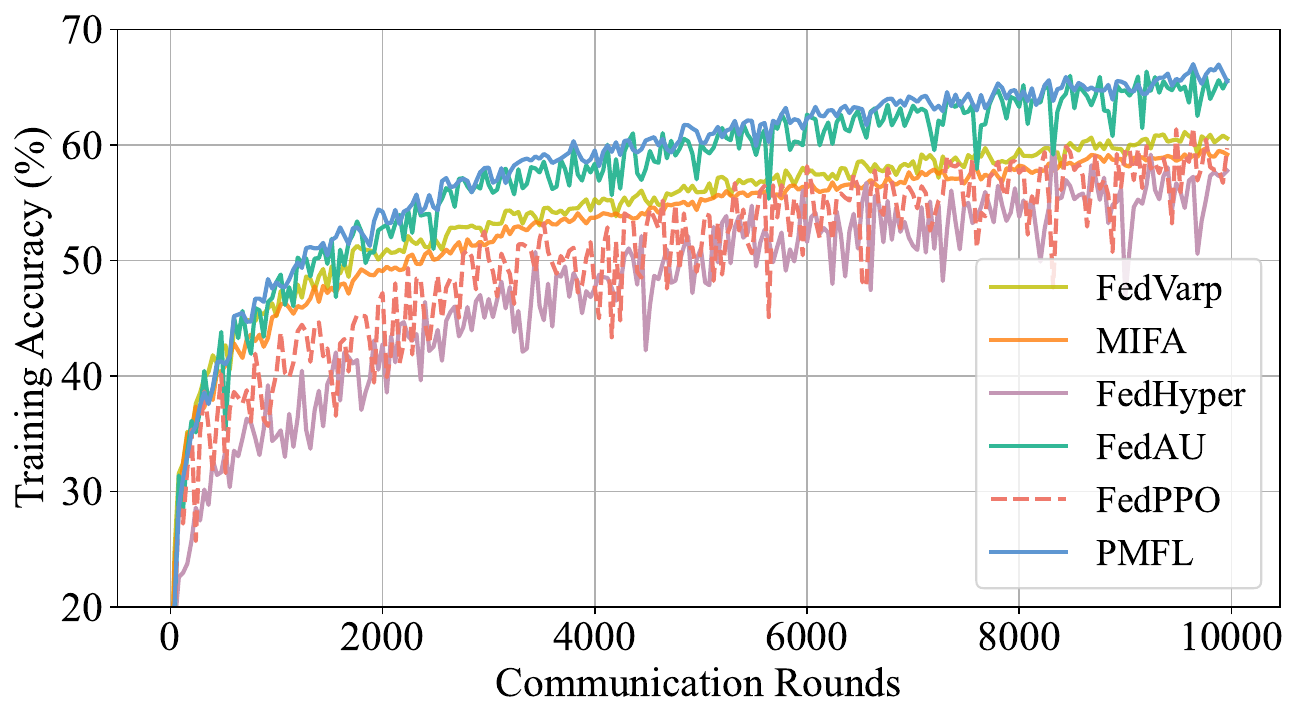}
  }
  \caption{Training accuracy of PMFL and the compared methods across different training rounds.}\label{3241058}
\end{figure}

To evaluate the convergence of different methods, we plot the training accuracy curves across multiple rounds for Bernoulli and Markovian patterns  on   SVHN, CIFAR10, and CINIC.
The results for PMFL and the baseline methods are shown in Fig. \ref{3241058}.
The results reveal that the training accuracy curves for FedVarp and MIFA exhibit extremely flat trajectories. 
This is because they incorporate the most recent updates from all nodes in each round, effectively implementing a ``full participation'' aggregation that produces smooth curves.
In contrast, the training curves of FedPPO, FedHyper, and FedAU exhibit significant fluctuations, indicating unstable global model performance during training.
This is because, under low participation rates, very few nodes actually participate in aggregation, making the global model's performance more susceptible to fluctuations.
Notably, PMFL's training accuracy gradually stabilizes with increasing rounds, demonstrating its robustness under heterogeneous scenarios.

\begin{table*}[t]
\centering
\caption{Accuracy results (\%) in various methods with different data heterogeneity on SVHN.}
      \renewcommand{\arraystretch}{1.15}
\scalebox{0.68}{
\tabcolsep = 0.37cm
\begin{tabular}{c|c|c|c|c|c|c|c|c|c|c|c|c|c|c|c}
\hline\hline
\multirow{3}{*}{\makecell{ \textbf{Participation}\\  \textbf{pattern}}} &\multirow{3}{*}{\textbf{Methods}} & \multicolumn{2}{c|}{$\alpha = 0.1$} & \multicolumn{2}{c|}{$\alpha = 0.3$} & \multicolumn{2}{c|}{$\alpha = 0.5$} & \multicolumn{2}{c|}{$\alpha = 0.8$} & \multicolumn{2}{c|}{$\alpha = 1.0$} &\multicolumn{2}{c|}{$\alpha = 2.0$}&\multicolumn{2}{c}{$\alpha_p = 5.0$}\\ \cmidrule{3-16}
&           & Training            & Test            & Training            & Test            & Training            & Test            & Training            & Test            & Training            & Test           &Training            & Test &Training            & Test \\ \midrule
\multirow{6}{*}{Bernoulli}&PMFL (ours) & \textbf{88.03}     & \textbf{87.47}  & \textbf{90.40}    & \textbf{89.88}   & \textbf{91.68}    & \textbf{90.95}    & \textbf{91.82}    & \textbf{91.16}   & \textbf{91.37}   & \textbf{90.65} &\textbf{92.51} &\textbf{91.60} &\textbf{91.79}&\textbf{91.15} \\
&FedVarp  \cite{jhunjhunwala2022fedvarp} & 85.72    & 85.45   & 89.06    & 88.69  & 90.26   & 89.63  & 89.96    & 89.42   & 89.93   & 89.34 & 90.98 &90.12 & 90.71&90.27\\
&MIFA  \cite{gu2021fast}   & 86.45     & 86.09    & 88.68    & 88.00    & 89.83   & 89.24   & 89.65    & 88.94   & 89.18    & 88.74  &90.57 &89.62 & 90.31&89.84 \\
&FedHyper  \cite{wang2023fedhyper}       & 84.90  & 83.95   & 89.19    & 88.81  & 89.71   & 89.30   & 90.74     & 90.02  & 90.36   & 89.74  &91.31 & 90.45 &91.17 &90.64 \\
&FedAU     \cite{wang2023lightweight}     & \underline{87.03}   & \underline{86.89}    & \underline{89.82}     & \underline{89.43}   & \underline{90.59}    & \underline{90.09}    & \underline{90.88}   & \underline{90.56}   & 90.76  & 90.23&\underline{91.71} &\underline{90.92}&\underline{91.46}&\underline{90.76}\\
&FedPPO  \cite{10909702}& 86.95    & 86.61    & 89.34   & 89.11  & 89.46   & 89.43   & 90.18   & 89.50  & \underline{90.86}   & \underline{90.34} & 91.17& 90.47&91.41&90.70\\\midrule
\multirow{6}{*}{Markovian}&PMFL (ours) &  \textbf{88.56}    &  \textbf{87.78}   &  \textbf{90.53}    &  \textbf{90.08}   &  \textbf{91.95}    &  \textbf{91.66}    &  \textbf{91.97}   &  \textbf{91.32}   &  \textbf{91.43}   &  \textbf{90.56} & \textbf{92.53} & \textbf{91.71} & \textbf{92.04}& \textbf{91.48} \\
&FedVarp  \cite{jhunjhunwala2022fedvarp} & 86.73  & 86.25   & 88.96    & 88.50   & 89.49   & 88.78   & 89.56   & 89.12   & 89.99   & 89.48 & 90.79 &90.12 & 90.42&89.90\\
&MIFA  \cite{gu2021fast}   & 86.86    & 86.48   & 88.34    & 87.97    & 89.19    & 88.54     & 88.77     & 88.35    & 88.63  & 88.15 &90.48 &89.71 & 89.39&88.87\\ 
&FedHyper    \cite{wang2023fedhyper}    & 84.34   & 82.96    & 88.85   & 88.17    & 89.92     & 89.32   & 90.25    & 89.87   &  \underline{90.67}    &  \underline{90.17}  &90.88& 90.26 &91.36 &90.91 \\
&FedAU   \cite{wang2023lightweight}       &  \underline{87.64}    &  \underline{87.01}   &  \underline{90.17}    &  \underline{89.56}    &  \underline{90.82}    &  \underline{90.48}   &  \underline{91.20}    &  \underline{90.59}  &  90.66   &  90.04 &  \underline{91.67} & \underline{90.99}& \underline{91.46}& \underline{90.83}\\
&FedPPO  \cite{10909702}& 86.25   & 85.84     & 89.55   & 88.83    & 90.07    & 89.72     & 90.75    & 89.79   & 90.40   & 90.07  & 90.99 & 90.10&91.38&90.52\\\midrule
\multirow{6}{*}{Cyclic}&PMFL (ours) & \textbf{87.90}    & \textbf{87.55}    & \textbf{90.63}   & \textbf{89.81}   & \textbf{91.30}   & \textbf{90.57}     & \textbf{91.83}    & \textbf{91.00}    & \textbf{91.10}   & \textbf{90.67} &\textbf{92.28} &\textbf{91.43} &\underline{91.66}&\underline{91.01}\\
&FedVarp  \cite{jhunjhunwala2022fedvarp} & 81.77    & 81.34    & 85.91    & 85.39   & 87.62     & 87.03    & 86.94   & 86.57    & 86.51    & 85.89  & 88.70 &88.46 & 87.76&87.42\\
&MIFA   \cite{gu2021fast}  & 76.13   & 75.65  & 77.75    & 77.47  & 86.23   & 85.77     & 84.90    & 84.17   & 76.63    & 76.35  &87.66 &87.54 & 83.35&82.78 \\
&FedHyper   \cite{wang2023fedhyper}      & 84.89     & 83.81   & 89.03   & 88.41    & 90.30   & 89.73   & 89.89   & 89.12   & 89.90  & 89.41 &90.65 & 89.97 &90.85 &90.38 \\
&FedAU   \cite{wang2023lightweight}& \underline{87.15}    & \underline{87.05}    & \underline{89.94}    & \underline{89.30}   & \underline{90.72}&\underline{90.15} & \underline{91.07}   & \underline{90.55}   & 90.41    & 89.98   & 91.38   & 90.79 &91.25 &90.56 \\
&FedPPO  \cite{10909702}& 86.96   & 86.72  & 89.68    & 89.25   & 90.89  & 90.13   & 90.92    & 90.34 &\underline{90.75}    & \underline{90.34}  & \underline{91.71} &\underline{90.86} &\textbf{91.73}&\textbf{91.14}\\ \hline\hline
\end{tabular}} \label{25522038}
\end{table*}

\subsubsection{The Impact of Data Heterogeneity for  Different Methods}
To evaluate the performance of PMFL and the compared  methods  under  varying degrees of data heterogeneity, we conduct  experiments on  SVHN. 
The data are split using a Dirichlet partition with concentration parameter $\alpha \in \{0.1, 0.3, 0.5, 0.8, 1.0, 2.0, 5.0\}$, where smaller $\alpha$   indicates higher data heterogeneity. 
The  results are shown in Table \ref{25522038}, where the best and second-best results highlighted in bold and underlined, respectively.
We observe that as data heterogeneity increases (i.e., as $\alpha$  decrease), the performance of all methods generally declines, and this trend is consistent across different participation patterns.
Notably, PMFL maintains high accuracy across all data heterogeneity conditions, which is attributed to its loss function that combines both classification term and model-contrastive term.
This design  minimizes local classification loss while enhancing the consistency of model updates across nodes, thereby mitigating the negative impact of data heterogeneity on the global model's performance.
Furthermore, we observe that MIFA performs poorly in the cyclic  pattern, regardless of the degree of data heterogeneity.
This is because each node alternates between participation and non-participation in fixed cycles, which leads to delayed historical model updates.
In summary,  the  results show that PMFL is more robust than the compared methods  under diverse data heterogeneity settings.

\subsubsection{Effects of Participation Heterogeneity Across Methods}
\begin{table*}[htbp]
\centering
\caption{Accuracy results (\%) in various methods on CIFAR10  with different Bernoulli participation heterogeneity.}
      \renewcommand{\arraystretch}{1.15}
\scalebox{0.7}{
\tabcolsep = 0.43cm
\begin{tabular}{c|c|c|c|c|c|c|c|c|c|c|c|c|c|c}
\hline\hline
\multirow{2}{*}{\textbf{Methods}}  & \multicolumn{2}{c|}{$\beta = 0.01$} & \multicolumn{2}{c|}{$\beta = 0.05$} & \multicolumn{2}{c|}{$\beta = 0.1$} & \multicolumn{2}{c|}{$\beta = 0.5$} & \multicolumn{2}{c|}{$\beta = 1.0$} &\multicolumn{2}{c|}{$\beta = 2.0$}&\multicolumn{2}{c}{$\beta = 5.0$}\\ \cmidrule{2-15}
            & Training            & Test            & Training             & Test            & Training             & Test            & Training             & Test            & Training             & Test           &Training             & Test &Train            & Test  \\\midrule
PMFL (ours) & \textbf{86.37}    & \textbf{78.08}    & \textbf{86.25}    & \textbf{78.15}    & \textbf{86.99} & \textbf{78.48}     & \textbf{89.62}   & \textbf{78.80}   & \textbf{89.46} & \textbf{79.25}   &\textbf{89.72} &\textbf{79.48} &\textbf{90.00}&\textbf{79.21} \\ 
FedVarp  \cite{jhunjhunwala2022fedvarp} & 78.90     & 74.88    & 82.06    & 77.12   & 83.06    & 77.83   & 83.13     & 76.78    & 84.41    & 77.96  & 86.25 &78.20 & 84.51&77.58\\ 
MIFA  \cite{gu2021fast}  & 80.14      & 76.16     & 81.00     & 77.25     & 81.72    & 77.43     & 82.94     & 76.89     & 83.66    & 77.50   &83.86  &77.31 & 83.98 &77.36  \\ 
FedHyper   \cite{wang2023fedhyper}      & 76.69  & 69.34     & 78.89    & 71.18    & 81.54    & 73.03    & 84.66    & 76.27    & 84.72     & 76.81 &86.77 & 77.80 &87.03 &78.39 \\ 
FedAU      \cite{wang2023lightweight}    & \underline{85.19}      & \underline{77.30}     & \underline{85.74}    & 77.44    & \underline{86.02}     & \underline{77.98}    & \underline{88.53}    & \underline{78.44}     & \underline{87.98}     & \underline{78.56}  & 88.50 &\underline{78.66} &\underline{88.67} &78.26 \\ 
FedPPO \cite{10909702}& 84.38    & 77.26     & 85.40     & \underline{77.48}    & 85.85  &77.02     & 86.76      & 78.04    & 87.02   & 78.28   & \underline{88.56}  & 78.62 &87.76 &\underline{78.62} \\ \hline\hline
\end{tabular}} \label{215413}
\end{table*}

To evaluate the performance of PMFL and the compared methods under varying levels of participation heterogeneity, we conduct experiments on  CIFAR10.
The participation heterogeneity is controlled by the parameter $\beta \in \{0.01, 0.05, 0.1, 0.5, 1.0, 2.0, 5.0\}$, where smaller $\beta$  indicates higher heterogeneity.
All experiments follow the Bernoulli  pattern.
The  results are shown in Table \ref{215413}.
Similar to the trend of data heterogeneity in Table \ref{25522038}, the performance of all methods generally degrades with increasing heterogeneity.
 In all methods,  PMFL demonstrates the best performance.
Furthermore,   under high participation heterogeneity (i.e., smaller $\beta$ values), the accuracy of MIFA and FedVarp outperforms that of FedHyper.
 However, as $\beta$ increases, their performance gradually drops below that of FedHyper. 
This phenomenon is due to MIFA and FedVarp both caching the last model update from each node for aggregation.
 When participation heterogeneity is high (i.e., smaller $\beta$), the server uses the last model update from each node to compute the global model, which  mitigates the impact of participation heterogeneity.
 In contrast, when node participation becomes  uniform (i.e., larger $\beta$), participation heterogeneity is alleviated.
 In this case, MIFA and FedVarp may aggregate stale last-submitted updates, which slows the convergence of the global model.

\subsubsection{Node-wise Distributions of Loss and Accuracy}

\begin{figure}[t!]
\centering
\subfloat[]
  {
      \label{111816121}  \includegraphics[width=0.48\linewidth]{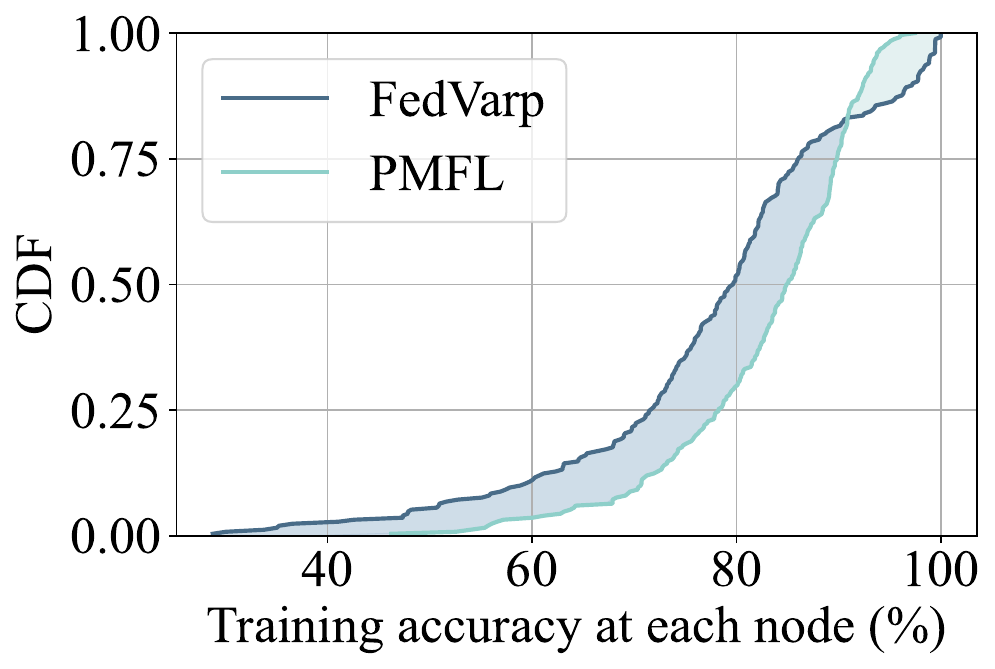}
  }
  \subfloat[]
  {
      \label{111816122}  \includegraphics[width=0.48\linewidth]{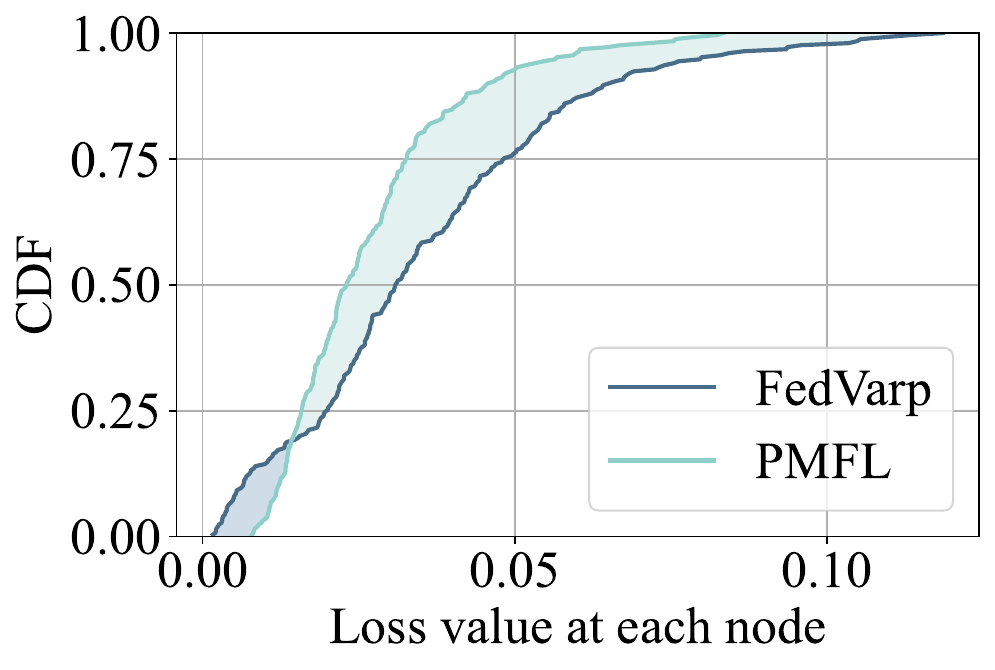}
  }
  \caption{Node-wise distributions of training accuracy and loss value on
SVHN  with Bernoulli participation.
}\label{1152236}
\end{figure}

To visualize  the  differences of the global model performance across nodes  in PMFL, we plot the Cumulative Distribution Functions (CDF) curves for accuracy and loss, and compare them with those of FedVarp.
The CDF describes the proportion of samples whose values are less than or equal to a given threshold, and thus reflects the distribution of a metric (e.g., accuracy or loss).
In the experiments,  we evaluate  the training accuracy and training loss of the trained global model at each node on SVHN and plot the CDF curves of these values, as shown in Fig. \ref{1152236}. 
Notably, the global model is trained under the Bernoulli  pattern.
We observe that, for PMFL, most nodes achieve training accuracy between 80\% and 100\%, and their training loss values are largely concentrated below 0.05.
In contrast, FedVarp exhibits a more dispersed distribution for both training accuracy and training loss. 
These results indicate that PMFL achieves more consistent model performance across nodes under heterogeneous conditions.

\subsubsection{Node Aggregation Weights in PMFL}

\begin{figure}[t]
\centering
\subfloat[]
  {
      \label{7051353}  \includegraphics[width=0.31\linewidth]{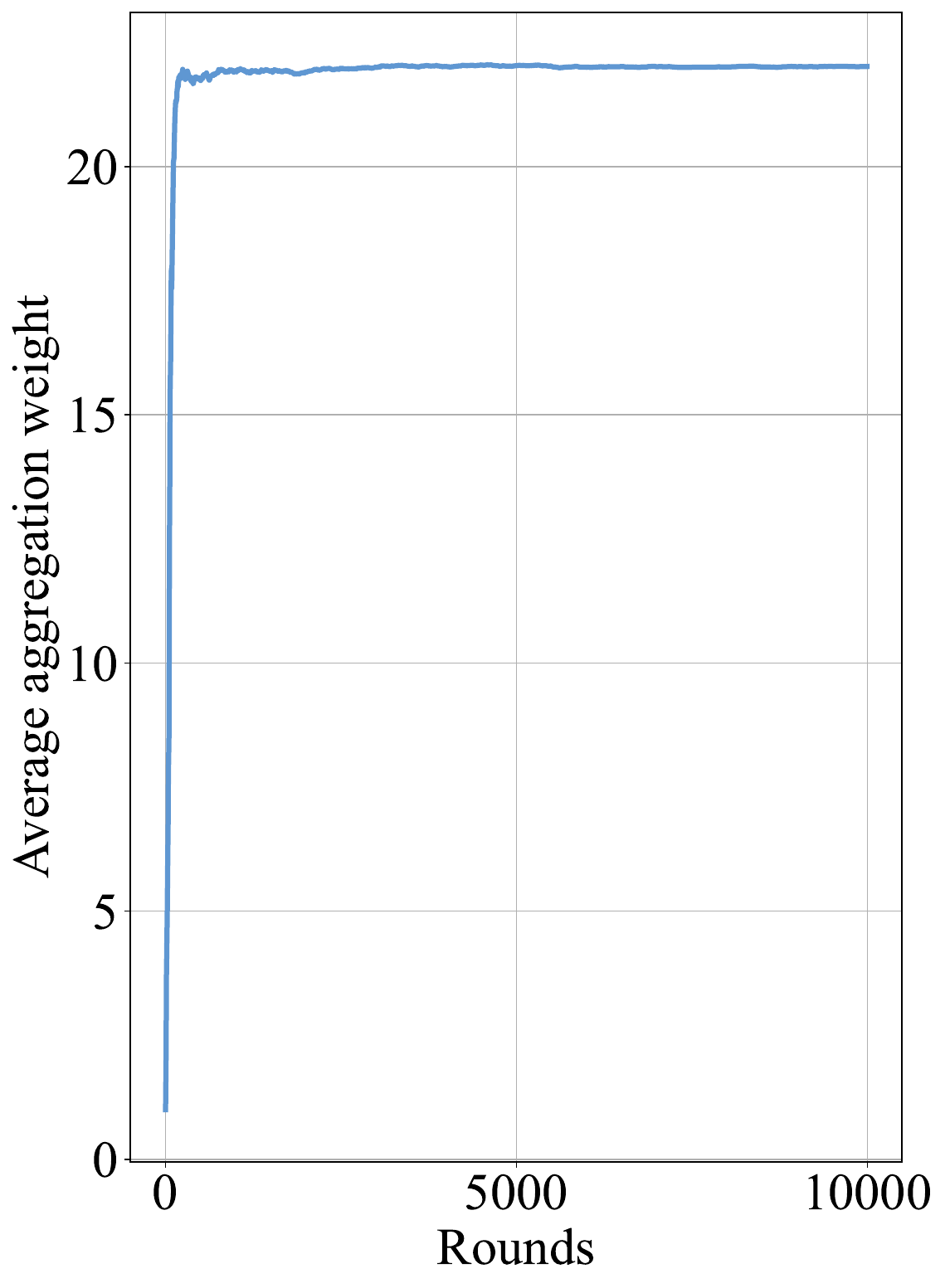}
  }
    \subfloat[]
  {
      \label{7051355}  \includegraphics[width=0.312\linewidth]{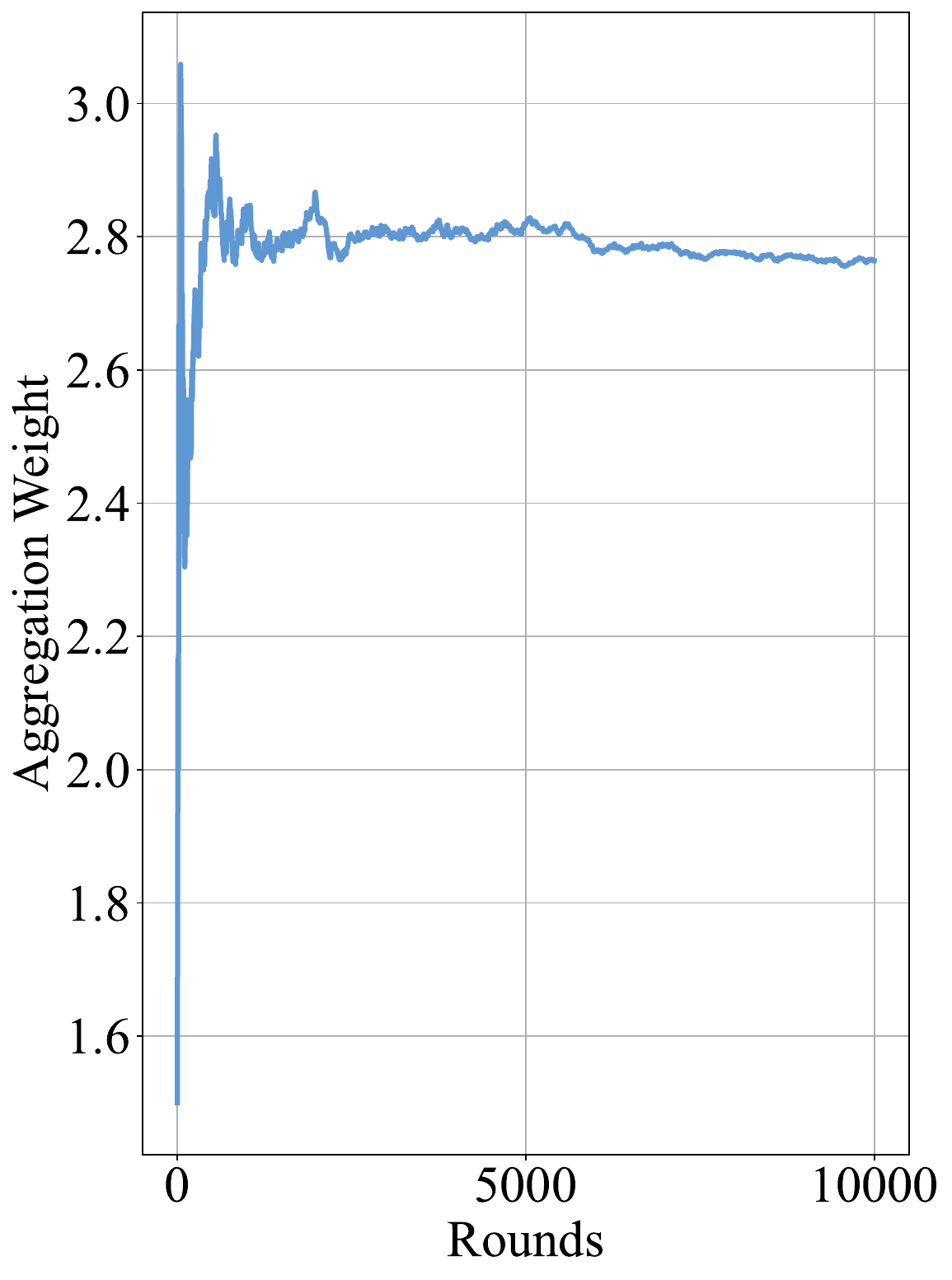}
  }
    \subfloat[]
  {
      \label{7051354}  \includegraphics[width=0.31\linewidth]{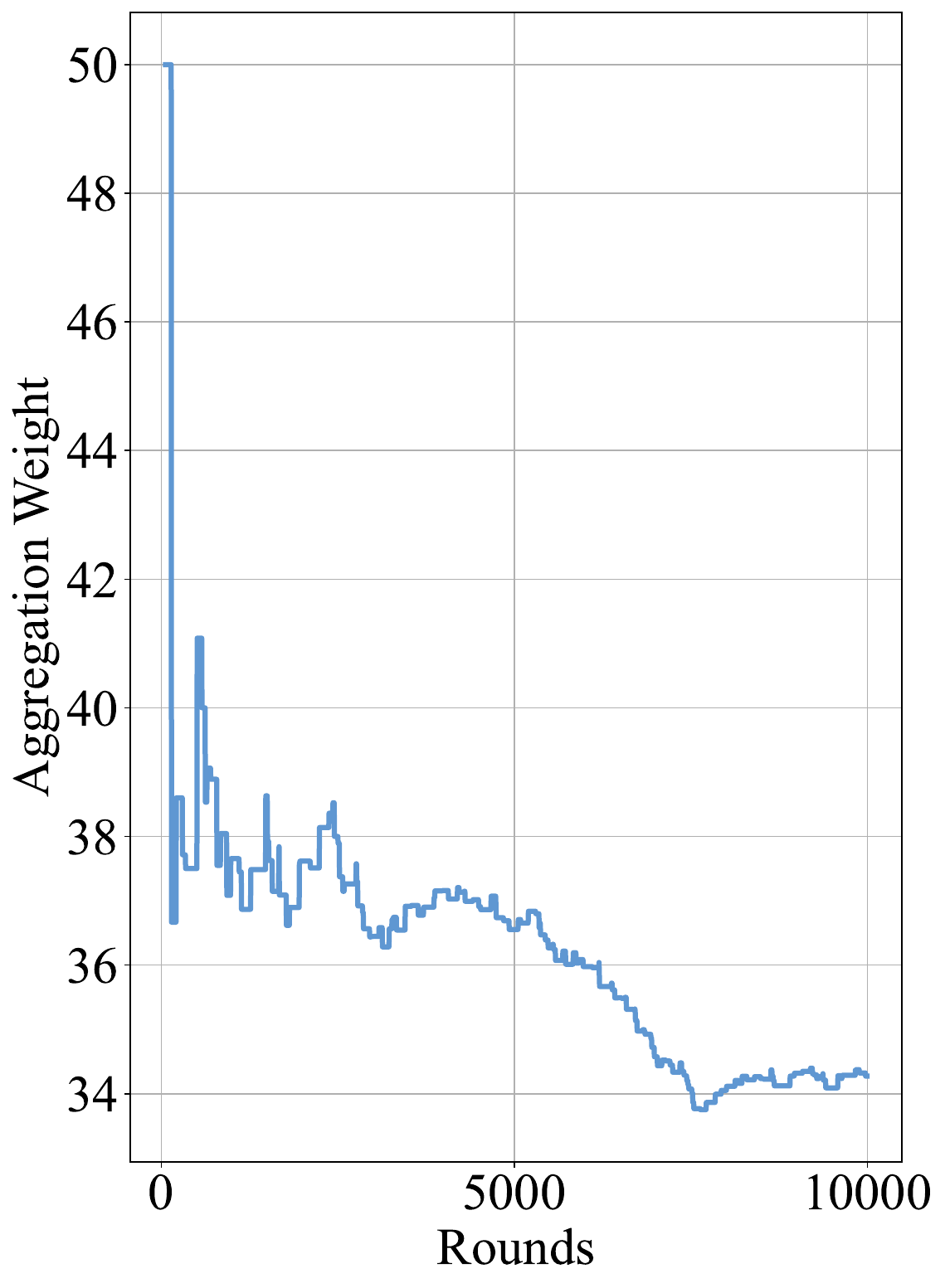}
  }
  \caption{Aggregation weights with Bernoulli participation, (a) average aggregation weights over all nodes, (b) aggregation weights of a single node with a high  participation frequency of 40.2\%, (c) aggregation weights of  a single node with a low  participation frequency of 2\%.
}\label{70513557}
\end{figure}

To explore the variation of node aggregation weights in PMFL, we plot the curve of node aggregation weights across communication rounds. 
In the experiments, we use  the CIFAR10  under the bernoulli pattern, with all parameters set to their default values. 
As shown in Fig. \ref{7051353}, we observe that the average aggregation weight across all nodes is approximately 22. 
As shown in Fig. \ref{7051355}, when a node has a high participation frequency (e.g., 40.2\%), its aggregation weight stays  above  2.8.
Conversely,  when the node participation frequency is low (i.e., 2\%), as shown in Fig. \ref{7051354}, its aggregation weight remains  above 30. 
This dynamic adjustment of aggregation weights effectively prevents the global model from being biased towards nodes with higher participation frequency. 
These  results validate that PMFL can automatically adjust aggregation weights according to each node's average participation interval.

\subsubsection{The Choice of Different Cutoff Interval $C$ on PMFL}
The cutoff interval $C$ affects each node's aggregation weight, and thus influences the performance of PMFL.
To evaluate the impact of $C$, we set $C \in \{20, 50, 100, 200\}$  to evaluate the performance of PMFL on the SVHN   under different participation patterns.
 All other parameters are set to their default values.
For each subtable, the best and second-best results are highlighted in bold and underlined, respectively.
The  results are reported in Table \ref{6181656}.
We observe that when $C$=200, the performance of PMFL decreases significantly.
This is because an excessively large $C$ amplifies  the aggregation weights of nodes that have not participated in training for many rounds, thereby hindering the convergence of the global model.
Conversely, if $C$ is set too small,  it may distort the actual participation frequencies of nodes, causing the optimization to deviate from the intended FL objective.
These results indicate  that PMFL requires a properly cutoff interval $C$.
We notice that the global model achieves the best performance when $C$=50, which is considered an appropriate setting.

\begin{table}[t]
\centering
\caption{Accuracy (\%) on training and test data of PMFL on SVHN, with different cutoff intervals $C$.}
\renewcommand{\arraystretch}{1.15}
\scalebox{0.7}{
\tabcolsep = 0.36cm
\begin{tabular}{c|c|c|c|c|c|c|c}
\hline\hline
\multirow{2}{*}{\makecell{ \textbf{Participation}\\  \textbf{pattern}}} & \multirow{2}{*}{$C$} & \multicolumn{2}{c|}{$\beta = 0.1$} & \multicolumn{2}{c|}{$\beta = 0.2$} & \multicolumn{2}{c}{$\beta = 0.4$} \\
        &    & Training            & Test            & Training            & Test            & Training            & Test          \\ \hline
   \multirow{4}{*}{Bernoulli}      & 20   & \underline{87.74}            & \underline{87.16}           & \textbf{87.89}           & \textbf{87.19}         & \underline{88.36}      & \underline{87.93}        \\
          &  50  & \textbf{88.03}          & \textbf{87.47}         & \underline{87.82}      & \underline{87.08}        & \textbf{88.87}         & \textbf{88.37}        \\
           &  100  & 87.22       & 86.51       & 85.48           & 84.52         & 87.85           & 87.59         \\
            &  200  & 84.56            & 83.84           & 84.18          & 83.70           & 87.29         & 87.11          \\ \hline
       \multirow{4}{*}{Markovian}         &  20  & \underline{87.83}           & \underline{87.20}           & \underline{87.64}           & \underline{87.01}           & \underline{88.53}          & \underline{88.28}         \\
              &  50  & \textbf{88.56}           & \textbf{87.78}           & \textbf{88.41}          & \textbf{87.67}           & \textbf{89.53}          & \textbf{88.95}        \\
               &  100  & 87.02         & 86.36         & 85.92           & 85.74           & 88.22        & 87.67        \\
                &  200  & 84.87      & 84.45       & 85.17       & 84.58           & 88.34            & 87.89           \\ \hline
        \multirow{4}{*}{Cyclic}            &  20  & \underline{87.52}           & \underline{87.02}          & \underline{84.24}      & \underline{83.83}           & \underline{87.35}            & \underline{87.14}           \\
                  &  50  & \textbf{87.90}            & \textbf{87.55}            & \textbf{87.16}           & \textbf{86.46}          & \textbf{87.62}           & \textbf{87.46}       \\
                   & 100   & 85.22           & 84.76        & 84.55            & 84.01          & 87.16           & 86.21         \\
                    & 200   & 81.37             & 80.62            & 81.06            & 80.30            & 86.81            & 85.74        \\ \hline\hline
\end{tabular}}\label{6181656}
\end{table}

\subsubsection{Ablation Study about PMFL}

\begin{table}[t]
\caption{Impact of each component in PMFL under CIFAR10 dataset.}
\tabcolsep= 0.22cm
\centering
\scalebox{0.9}{
\begin{tabular}{cccccc}
\hline\hline 
 \multirow{2}{*}{Methods} &  \multirow{2}{*}{\makecell{Data \\Distribution}}  &\multicolumn{4}{c}{Test Accuracy (\%)}  \\
&&$\beta = 0.01$ &$\beta = 0.05$  &$\beta = 0.10$ &$\beta = 0.50$ \\\hline 
 \multirow{2}{*}{PMFL} &$\alpha = 0.5$&79.04  &\textbf{79.27}  &\textbf{79.49} &79.73\\
&$\alpha = 0.1$&\textbf{78.08} &\textbf{78.15}  &\textbf{78.48} &\textbf{78.80}\\\hline 
 \multirow{2}{*}{w/o MCT} &$\alpha = 0.5$&78.94 &79.06 &79.23 &79.45\\
&$\alpha = 0.1$&77.63  &77.81  &78.26 &78.49\\\hline 
 \multirow{2}{*}{w/o AWC} &$\alpha = 0.5$&71.86 &73.05  &75.07&75.80\\
&$\alpha = 0.1$&69.72  &72.18  &74.72&75.35\\\hline 
 \multirow{2}{*}{w/o HGM} &$\alpha = 0.5$& \textbf{79.13} &79.24 &79.39 &\textbf{79.96}\\
&$\alpha = 0.1$&77.61  &77.73  &77.96 &78.25\\\hline 
\hline
\end{tabular}}
\justifying
Note: ``MCT'' represents the model-contrastive term, ``AWC'' denotes the adaptive aggregation weight calculation step, and ``HGM'' denotes the historical global models. Best rusults are in \textbf{blod}.
\label{3159191351}
\end{table}

By incorporating the model-contrastive term in local model training, designing the adaptive aggregation weight calculation step and introducing historical  information in during the aggregation phase, PMFL improves FL performance in heterogeneous environments.
To evaluate the  contributions of each component in PMFL, we conduct an ablation study. 
Specifically, we evaluate PMFL on the CIFAR10  under the bernoulli  pattern.
The results are reported in Table \ref{3159191351},  where ``w/o'' denotes ``without''.
We observe that removing the MCT (i.e., the model-contrastive term)  reduces the test accuracy of the global model.
This is because, without the model-contrastive term, local training relies solely on the classification term, which exacerbates inconsistency among  model updates.
Moreover, under different heterogeneous conditions, PMFL's performance degrades significantly when AWC is removed.
This is because the global optimization objective of FL shifts toward the local optimization objectives of nodes with higher participation frequency.
Notably, when the data heterogeneity is $\alpha$=0.5, removing HGM (i.e., historical global models) has only a minor impact on PMFL's test accuracy.
 However, when  $\alpha$=0.1, removing HGM decreases the test accuracy by more than 0.5\% compared with PMFL.
This indicates that integrating historical global models is particularly beneficial under highly heterogeneous data distributions.
In summary, the   results show that each component in PMFL contributes positively to improving the performance of FL in heterogeneous scenarios.

\subsubsection{Consistency of PMFL in Non-IID Settings}
The designed model-contrastive term promotes the alignment between local and global models, thereby improving the consistency of model updates across nodes.
To assess this improvement in consistency, we measure the deviation among model updates, which is defined as follows.

\begin{myDef}\label{11281514} (Deviation)
Given that in the $t$-th  round, each  node $k \in \mathcal{K}^t$ submits its  model update $\Delta_k^t$, the deviation among model updates  is given by:
\begin{equation}\scalebox{1}{$
\mathrm{Dev}^t = \sum_{k \in \mathcal{K}^t} (1-\operatorname{Sim}(\Delta_k^t, \bar{\Delta}^t)),\label{83121262}$}\nonumber
\end{equation} 
where $\operatorname{Sim}(\cdot)$ is the cosine similarity between two vectors, and $\mathrm{Dev}^t$ measures the  deviation among  model updates.  
\end{myDef}
\begin{Remark}
If all $k_{1}, k_{2} \in \mathcal{K}^t$ satisfy $\Delta_{k_{1}}^t = \Delta_{k_{2}}^t$, then the relation  $\bar{\Delta}^t = \Delta_k^t$ holds.
In this case, the model updates are perfectly consistent, which implies  $\mathrm{Dev}^t = 0$.
Conversely,   a larger $\mathrm{Dev}^t$ reflects significant inconsistency among updates.
\end{Remark}

\begin{figure}[t]
  \centering
  \subfloat[$\beta =0.1$]
  {
      \label{12251442}  \includegraphics[width=0.45\linewidth]{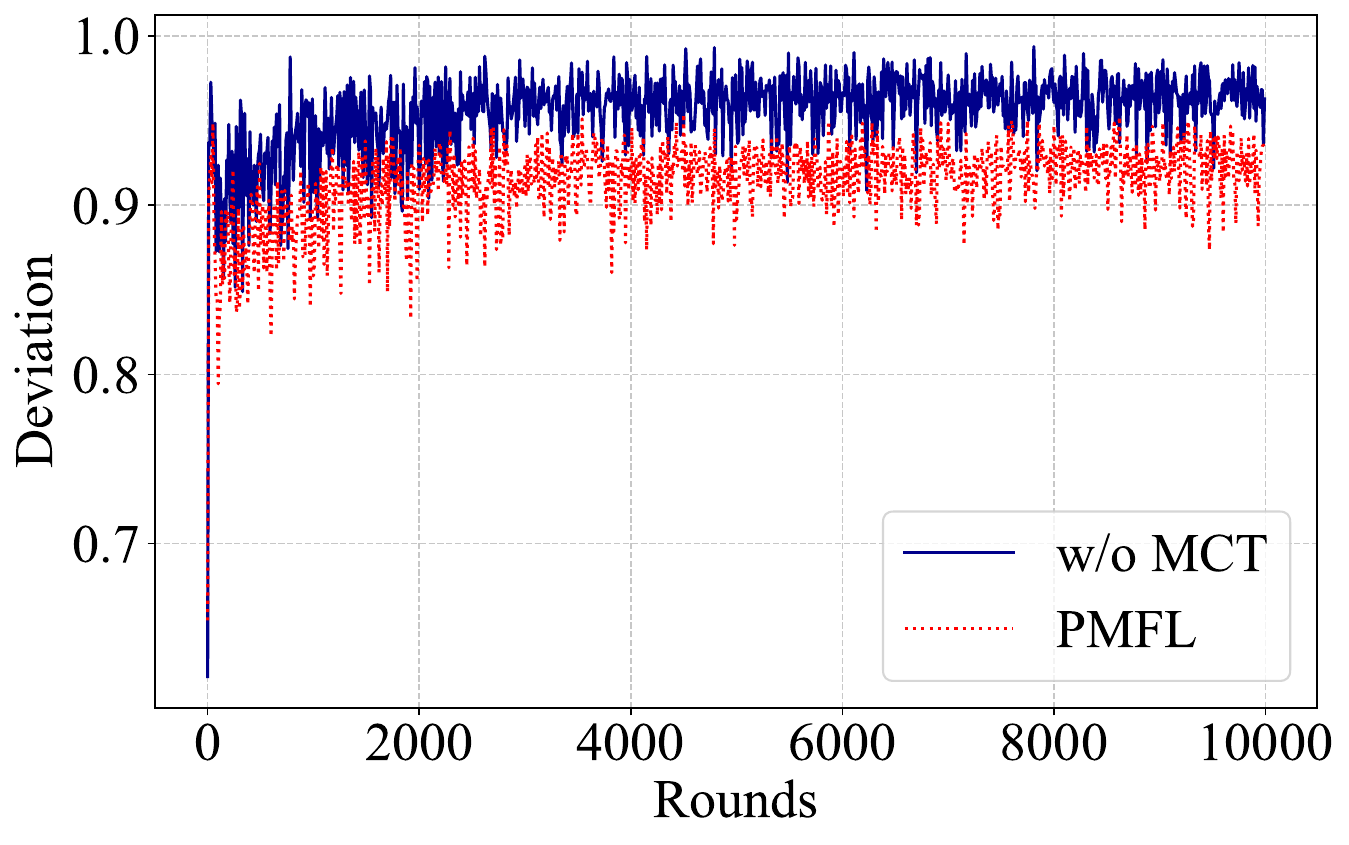}
  }
    \subfloat[$\beta=0.3$]
  {
      \label{12251443}  \includegraphics[width=0.45\linewidth]{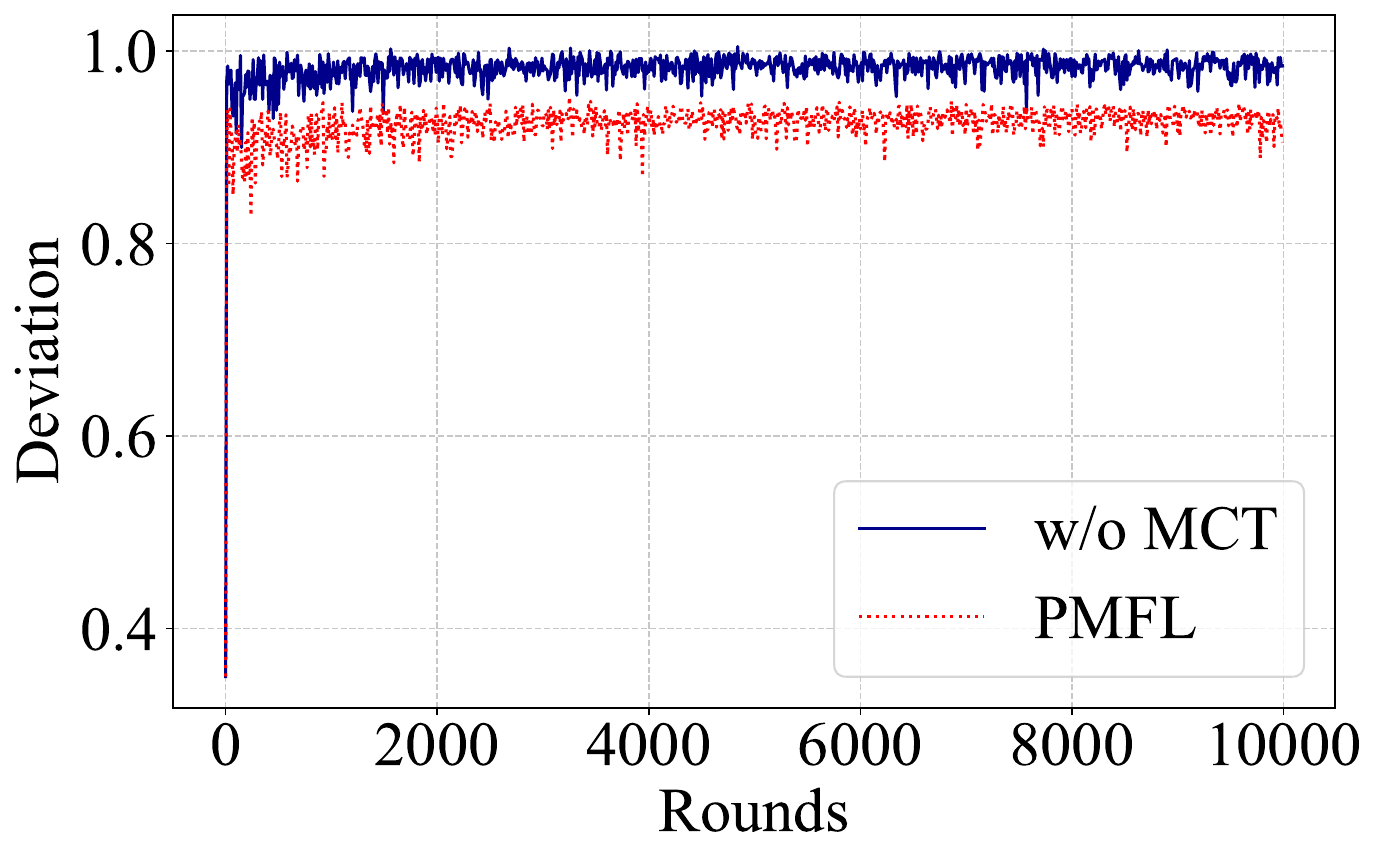}
  }
\caption{Deviation of PMFL under varying participation heterogeneity levels.}
\label{12251444}
\end{figure}

According to Definition \ref{11281514}, we measure the  deviation  among nodes' model updates in  PMFL. 
For comparison, we evaluate an ablated PMFL variant without the model-contrastive term in the local objective (w/o MCT).
Experiments are conducted on CIFAR10  under bernoulli pattern, and the results are reported in Fig. \ref{12251444}.
We observe that PMFL achieves a  lower   deviation than w/o MCT.
This reduction is mainly attributed to the model-contrastive term incorporated into PMFL's local objective, which strengthens the alignment between local and global models and thereby improves the consistency of model updates across nodes.
These results demonstrate the effectiveness of the proposed model-contrastive term.

\subsubsection{Impact of  Local and Global Sliding Buffer   on PMFL}
\begin{figure}[t]
  \centering
  \subfloat[Bernoulli]
  {
      \label{113020071}  \includegraphics[width=0.3\linewidth]{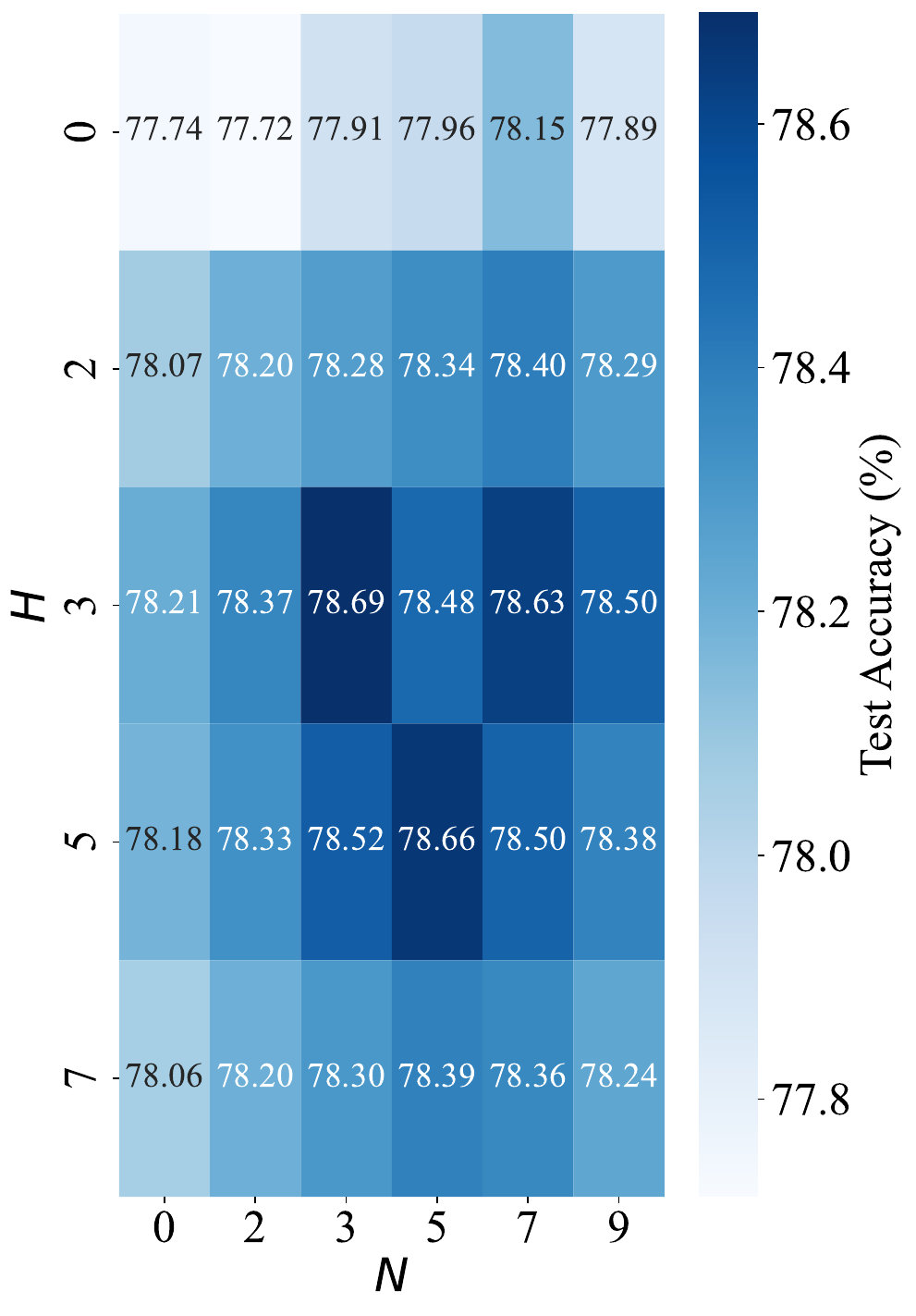}
  }
    \subfloat[Markovian]
  {
      \label{113020073}  \includegraphics[width=0.3\linewidth]{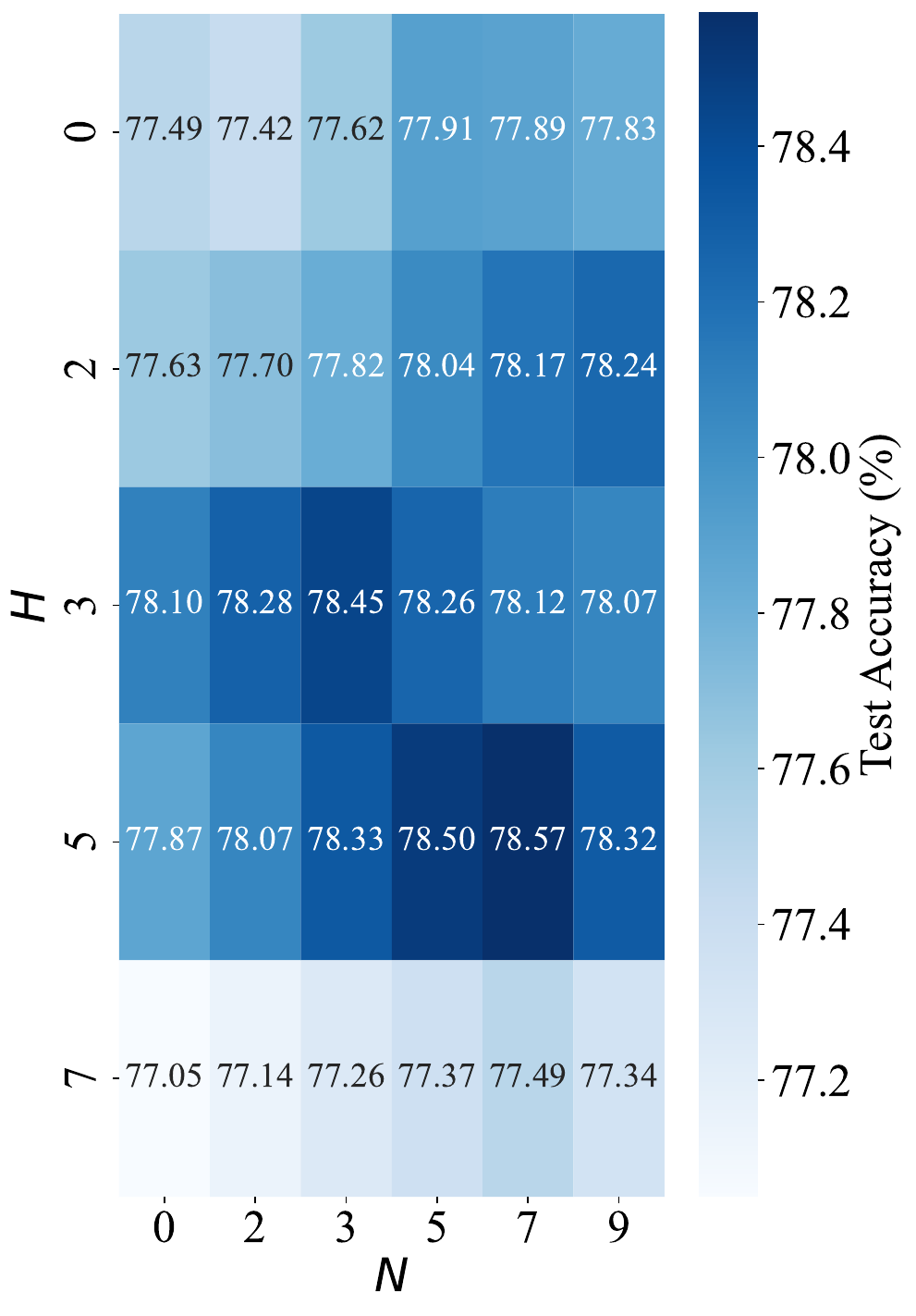}
  }
      \subfloat[Cyclic]
  {
      \label{113020075}  \includegraphics[width=0.3\linewidth]{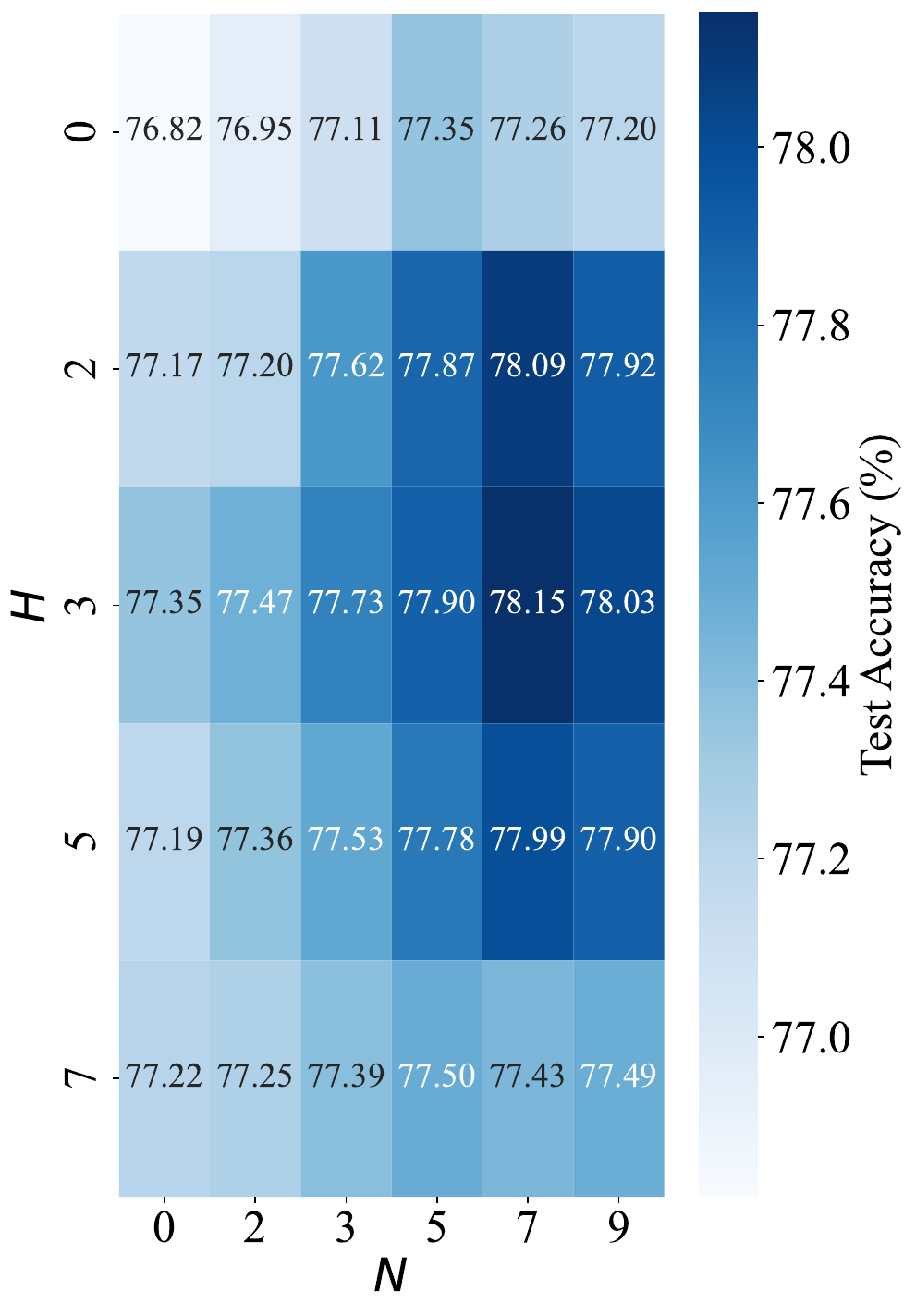}
  } 
\caption{Accuracy results of PMFL under   local    and global sliding buffer sizes on the CIFAR10.}
\label{11302007}
\end{figure}

Each node   in PMFL maintains a local sliding buffer to store its historical local models, thereby providing more positive and negative samples.
Meanwhile, during the aggregation phase, the server introduces a global sliding buffer to store historical global models, enabling the updated global model to integrate information from previous rounds.
To evaluate the sensitivity of PMFL to the local buffer size $N$ and global buffer size $H$, we conduct experiments on the CIFAR10, where $H$ is selected from the range   $\{0,2,3,5,7,9\}$, and  $N$ is selected from $\{0,2,3,5,7\}$, with other parameters kept at their default values.
The  results are shown in Fig. \ref{11302007}.
We observe that when $H$$\textgreater$0, PMFL outperforms the $H$=0 setting across  all participation  patterns, indicating that integrating historical global models into aggregation improves FL performance in heterogeneous scenarios.
However, when the global buffer size increases to $H$=7, the performance begins to decline.
Although a larger global buffer allows more historical global models to contribute to the latest global model, stale  information may  interfere with  the optimization process, hindering the FL convergence.
These results suggest that integrating historical global models is beneficial, but overreliance on outdated global models may degrade performance.
Additionally, we observe that increasing  the local buffer size enhances the performance of the global model.
A larger local buffer provides more positive and negative samples for computing the model-contrastive loss, which strengthens the effectiveness of the model-contrastive term.
Thus, properly configuring both local and global buffers plays a positive role in boosting the performance of PMFL.


\subsubsection{Global Model Performance Fluctuations of PMFL}

\begin{figure}[t]
  \centering
  \subfloat[]
  {
      \label{11302007113}  \includegraphics[width=0.45\linewidth]{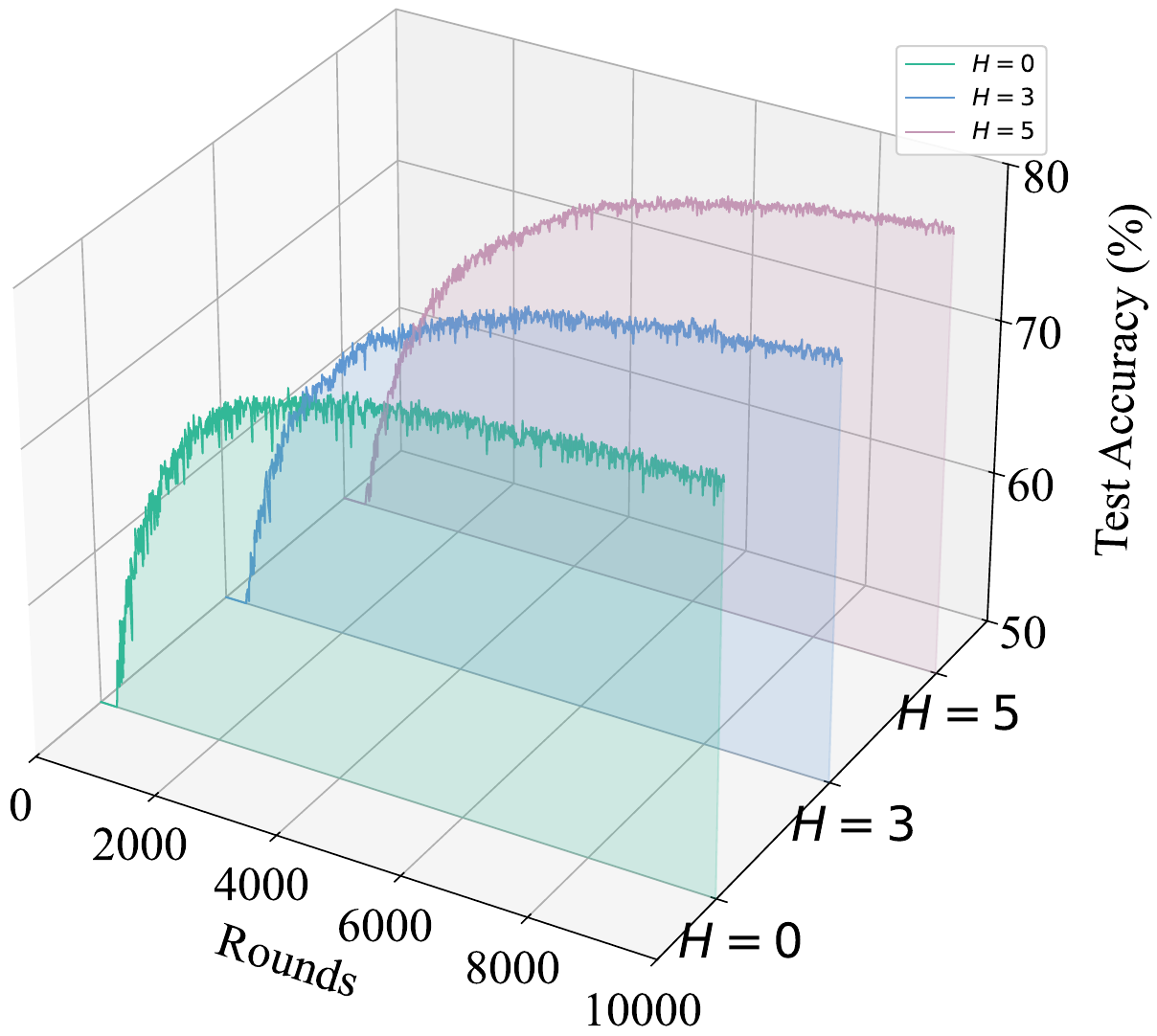}
  }
    \subfloat[]
  {
      \label{11302007345}  \includegraphics[width=0.45\linewidth]{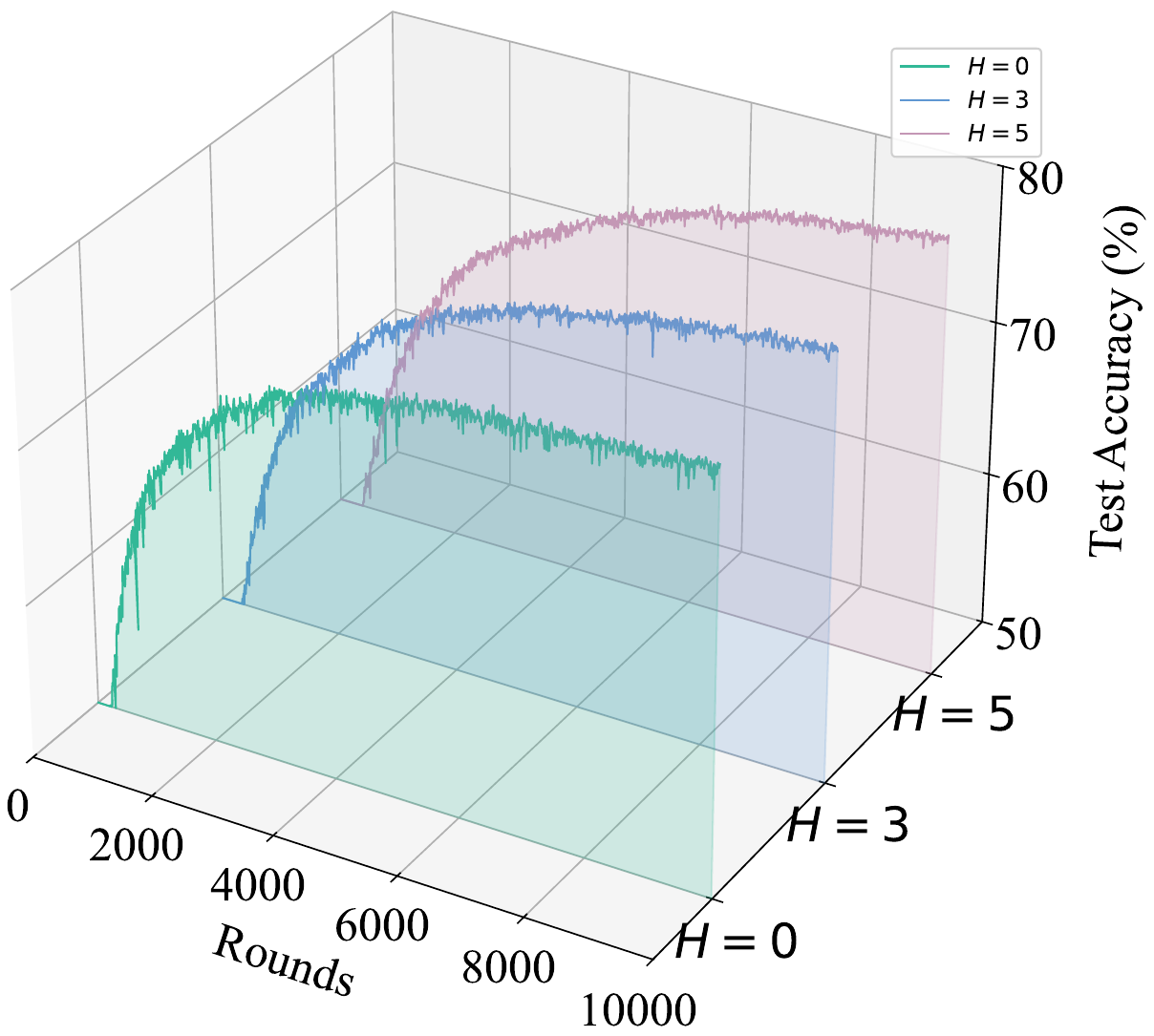}
  }
\caption{
Test accuracy of the global model in MCFL over  rounds under different expected participation frequencies $\mathbb{E}[p_k]$. (a) $\mathbb{E}[p_k]$=0.05,  (b) $\mathbb{E}[p_k]$=0.10.}
\label{11302007000}
\end{figure}

In PMFL, the historical global models are incorporated into the aggregation phase  to mitigate performance fluctuations of the global model across adjacent  rounds.
To evaluate  its effectiveness, we conduct experiments     on the CIFAR10  under the bernoulli  pattern.
The expected participation frequency $\mathbb{E}[p_k]$ is set to 0.05 and 0.10, respectively, and the global sliding buffer size $H$ is set to 0, 3, and 5.
The results are shown in Fig. \ref{11302007000}.
We observe that  the global model's test accuracy exhibits the largest fluctuations when when $H$=0.
 As the global sliding buffer size increases, the fluctuations gradually diminish.
 This indicates that PMFL effectively mitigates the instability of the global model under low participation rates by integrating historical global models.
 This facilitates more stable global representations for model-contrastive term.
However, Fig. \ref{11302007}  shows that an overly large global buffer   degrades the global model's performance. 
This is because incorporating too many historical models amplifies the smoothing effect, causing the global updates to rely excessively on outdated  information.

\subsubsection{Impact of Learning Rates $\eta_{l}$ and   $\eta_{g}$ on  PMFL}

\begin{figure}[t]
\centering
    \subfloat[CIFAR10]
      {\label{5302115114}
  \centering \includegraphics[width=0.31\linewidth]{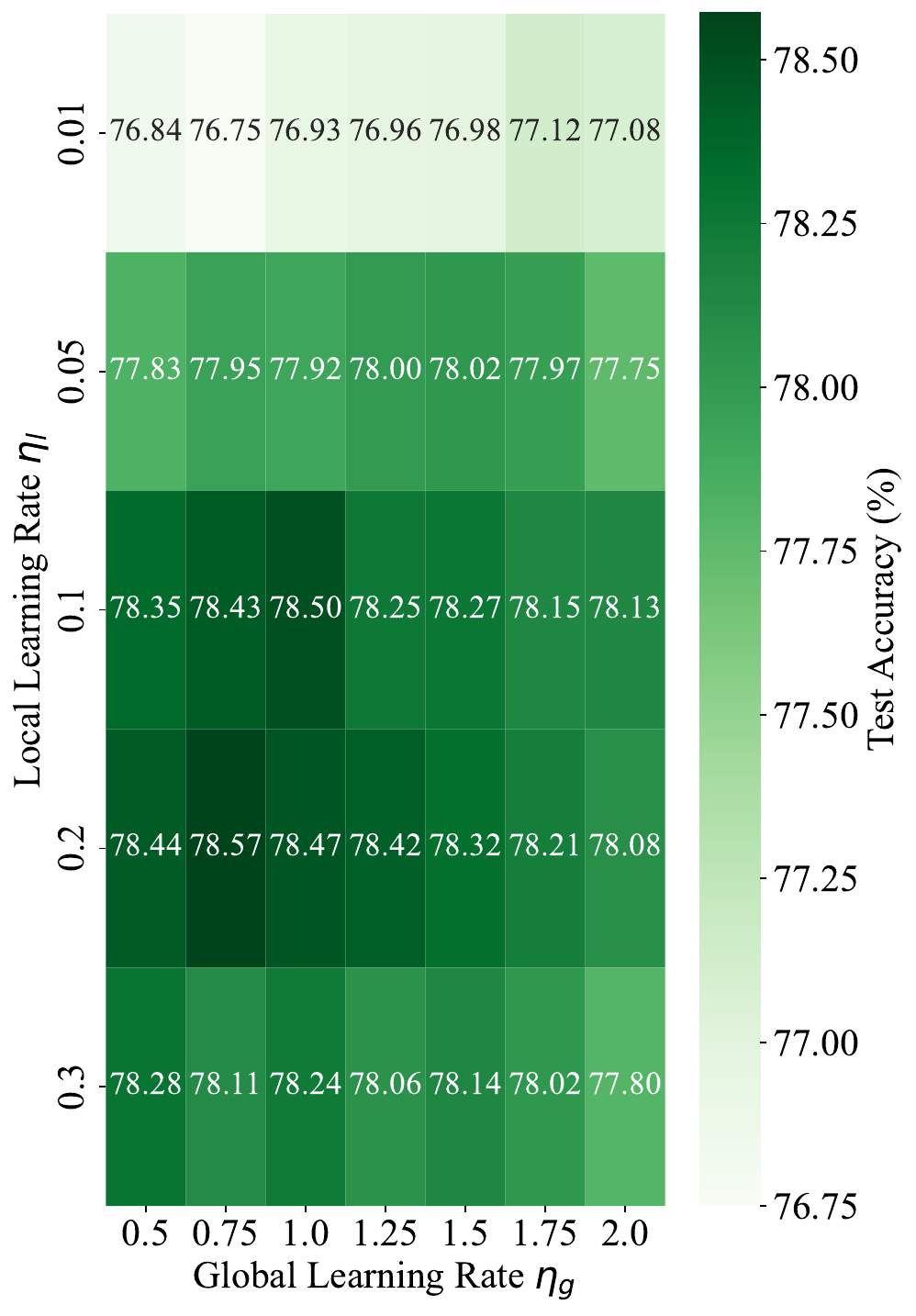}
  }
      \subfloat[CINIC]
      {\label{5302115213}
  \centering \includegraphics[width=0.31\linewidth]{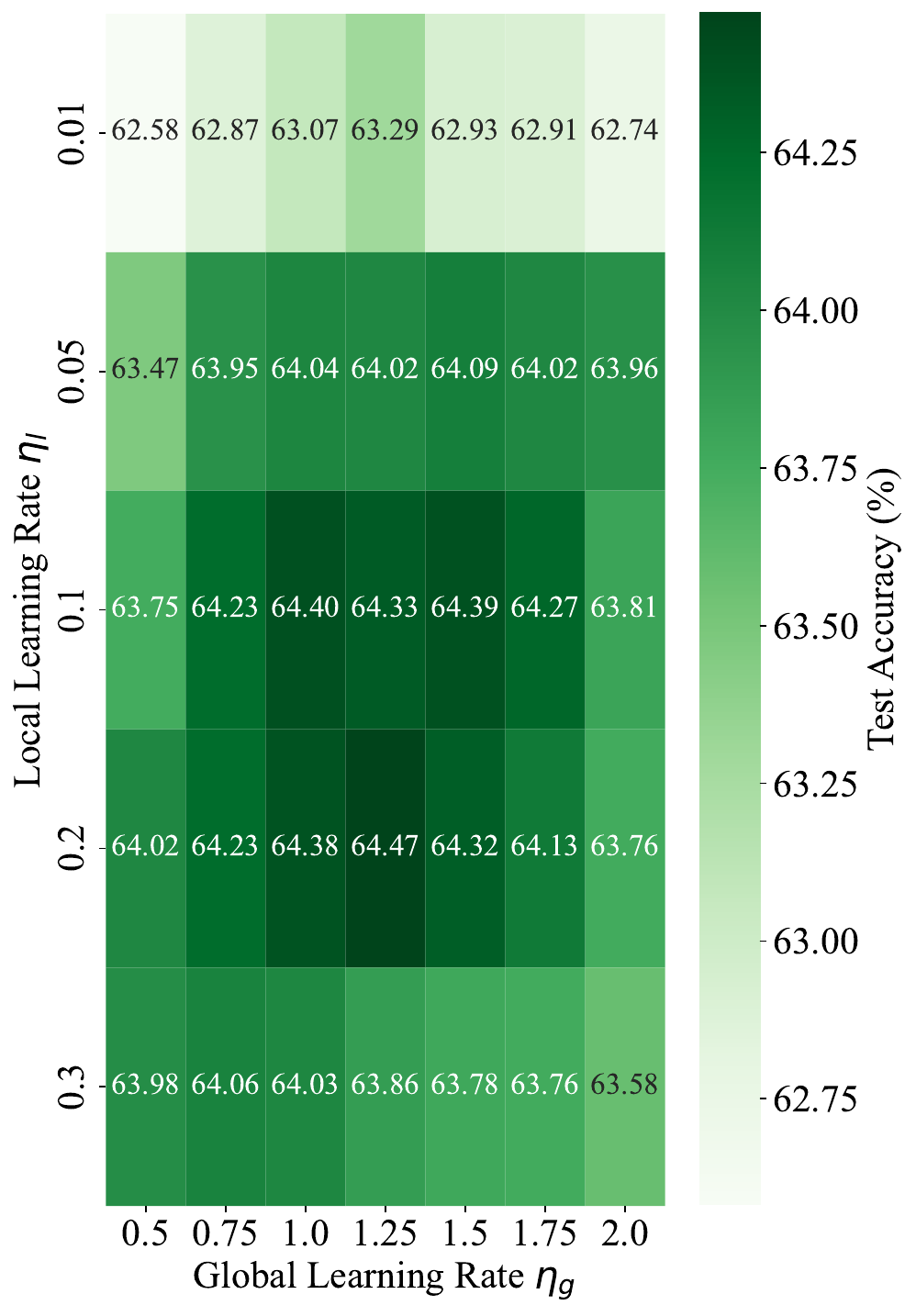}
  }
 \subfloat[CIFAR100]
      {\label{5302115212}
  \centering \includegraphics[width=0.31\linewidth]{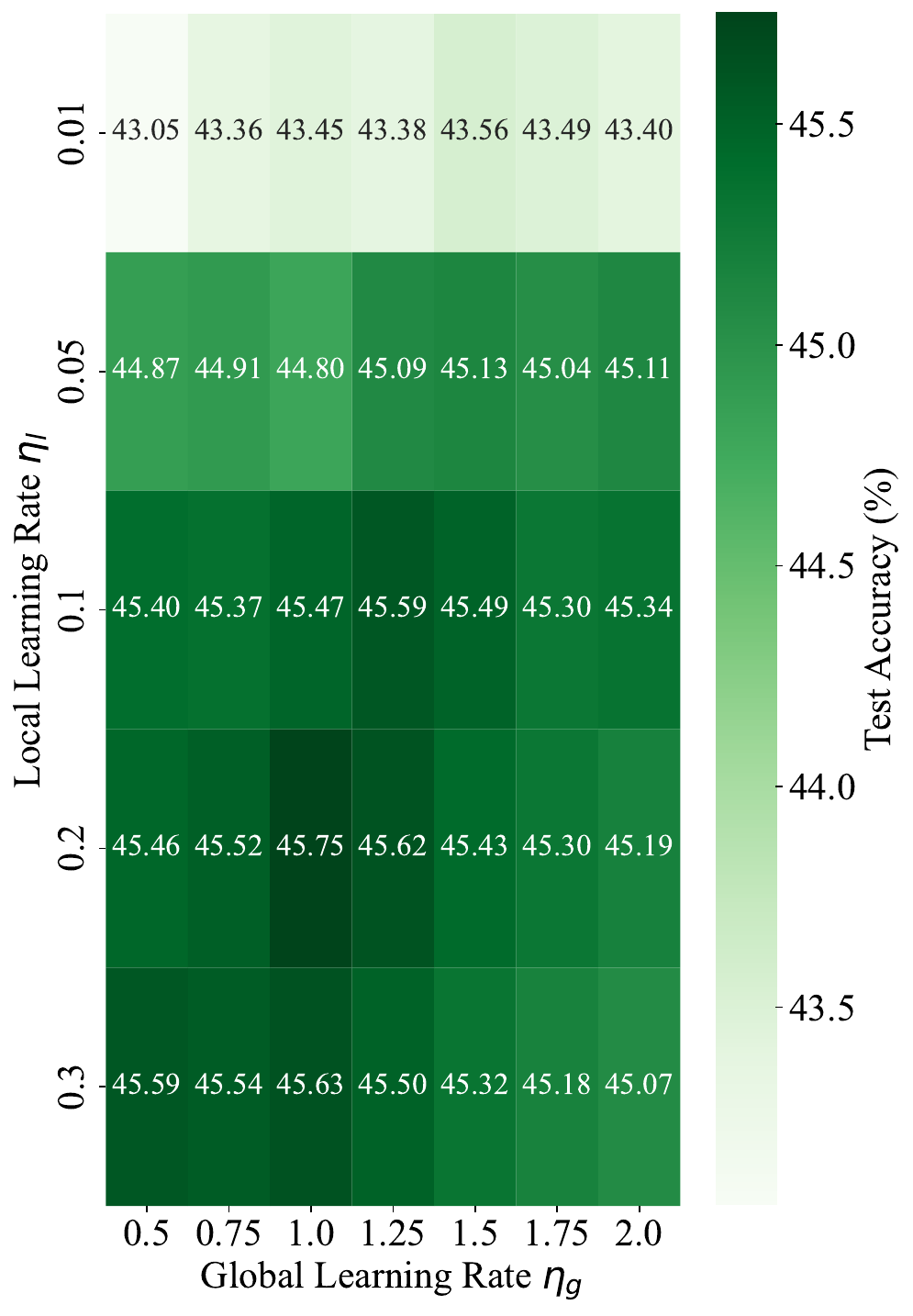}
  }
\caption{Test accuracy of PMFL with varying   learning rates $\eta_l$ and $\eta_g$.}
\label{530211541}
\end{figure}

Different local learning rates $\eta_l$ and global learning rates $\eta_g$ can affect the performance of  FL.
To evaluate the sensitivity of PMFL under different learning rate configurations, we conduct experiments on the CIFAR10, CINIC, and CIFAR100   under the Bernoulli   pattern.
Specifically, the global learning rate is selected from $\{0.5, 0.75, 1, 1.25, 1.5, 1.75, 2.0\}$, while the local learning rate is chosen from $\{0.01, 0.05, 0.1, 0.2, 0.3\}$.
The test accuracy results are presented in Fig. \ref{530211541}, where deeper green indicates higher accuracy.
The results show that different local learning rate settings lead to accuracy variations of 0\%-2\%, with PMFL achieving the lowest performance when the local learning rate is set to 0.01.
This is because, under a limited number of  rounds, smaller local learning rates reduce the magnitude of model updates, thereby slowing model convergence.
In addition, we observe that changes in the global learning rate have a relatively small impact on performance.
This is because PMFL's global update depends primarily on the  model updates uploaded by the nodes, whose update magnitudes are mainly determined by the local learning rate.
Overall, these results indicate that PMFL exhibits low sensitivity to both local and global learning rate settings.

\subsubsection{Impact of Model-Contrastive Term on PMFL}

\begin{figure}[t]
\centering
\subfloat[Training set]
  {
      \label{4310561}  \includegraphics[width=0.48\linewidth]{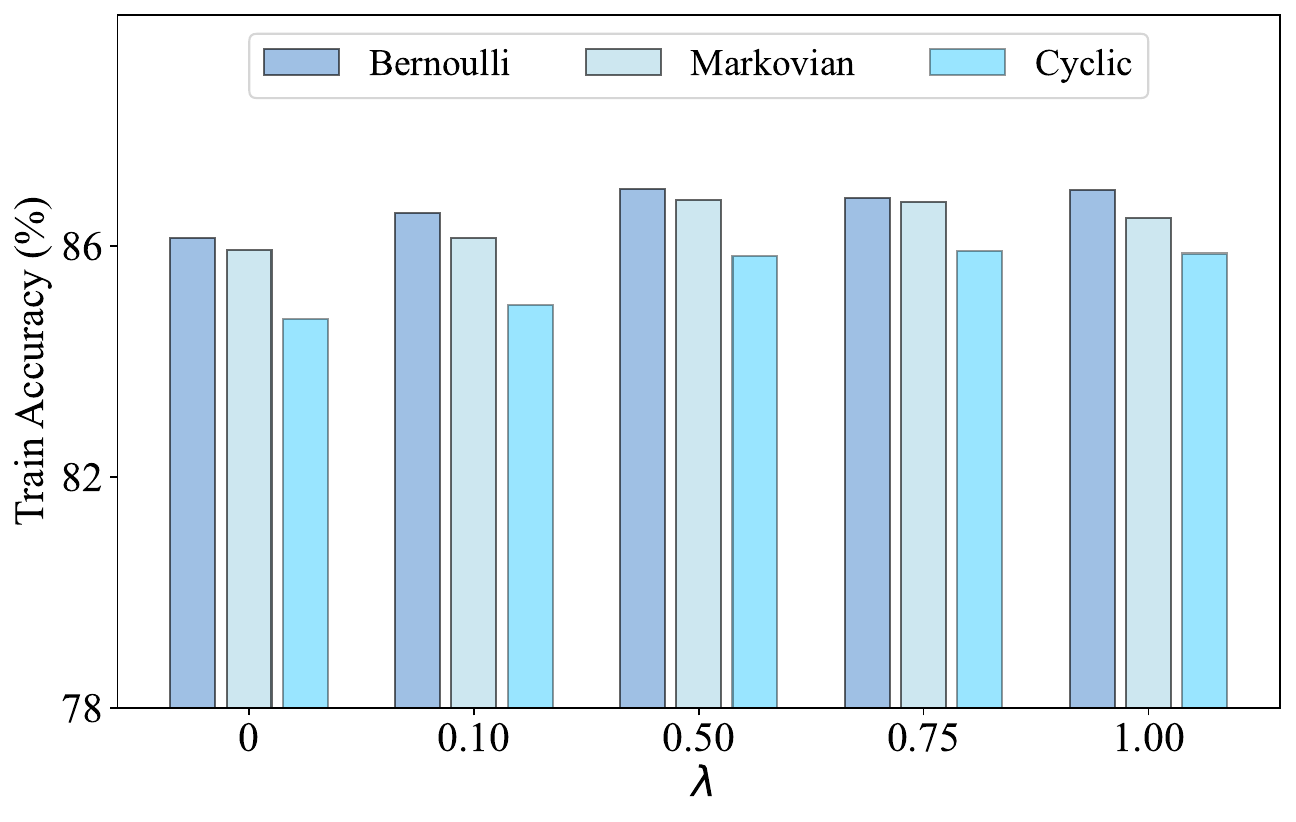}
  }
  \subfloat[Test set]
  {
      \label{4310562}  \includegraphics[width=0.48\linewidth]{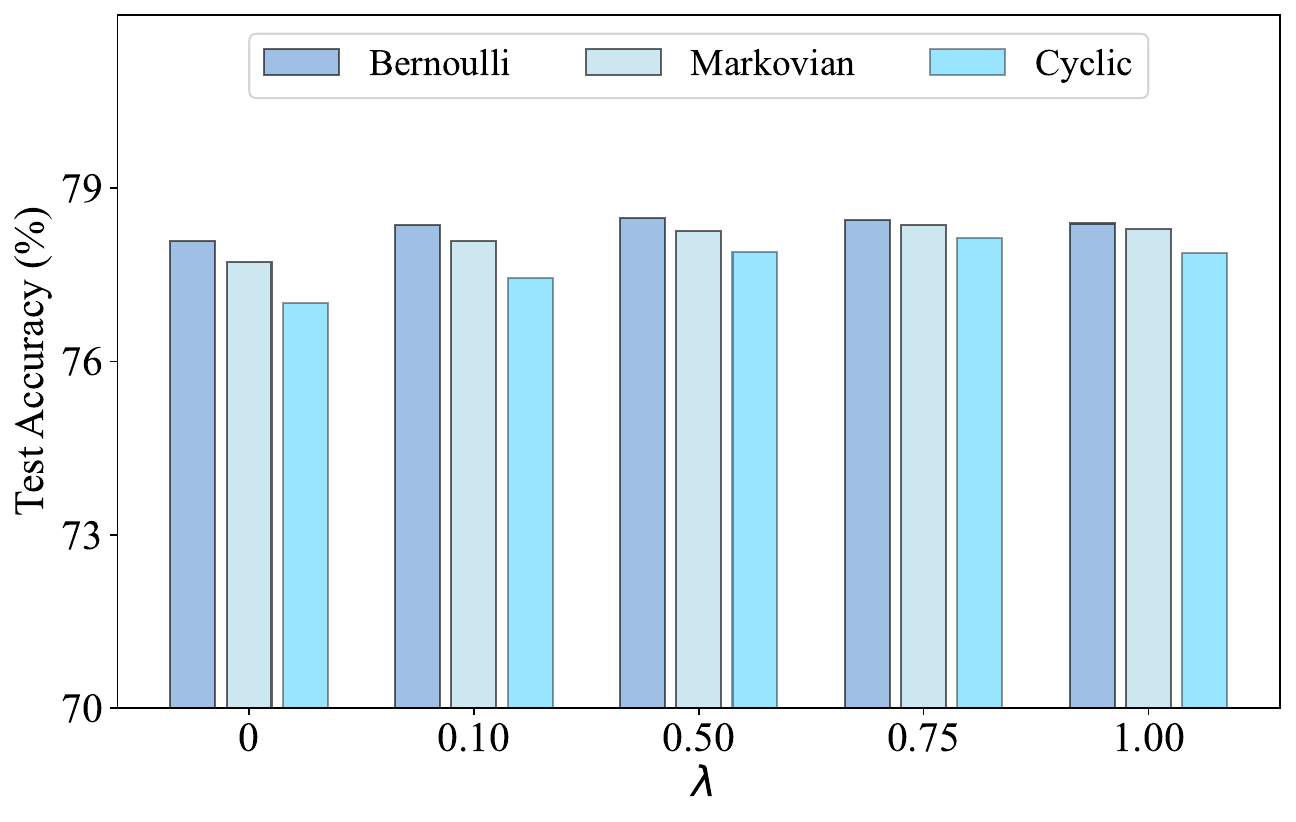}
  }
  \caption{Accuracy results  with varying  importance  coefficients   $\lambda$ in PMFL.
}\label{431056}
\end{figure}

To evaluate the impact of the importance coefficient $\lambda$ of the model-contrastive term on PMFL, we conduct experiments on  CIFAR10  under three participation patterns, with $\lambda \in \{0, 0.10, 0.50, 0.75, 1.00\}$.
Fig. \ref{431056} reports the performance of PMFL under different $\lambda$ values.
The results show that when $\lambda$=0, PMFL  achieves the worst performance across all participation patterns. This is because setting $\lambda$=0 removes the model-contrastive term from the optimization function,  weakening the consistency among model updates across nodes and degrading FL performance under data heterogeneity.
As $\lambda$ increases, the performance of PMFL gradually improves, indicating that the model-contrastive term   improves FL performance in heterogeneous data settings.
However, as $\lambda$ increases to 0.75 and 1.00, the performance of the PMFL reaches a stable state.
 This is because the performance gain provided by the model-contrastive term is  limited.
They do not continue to increase with larger values of $\lambda$.
Moreover, an excessively large $\lambda$ may cause ``over-alignment'' between local and global model representations,  thereby restricting the model's classification performance.
Thus, the importance coefficient $\lambda$ should not be set excessively large.

\subsubsection{Stability of PMFL}

\begin{figure}[t]
\centering
    \subfloat[CIFAR10]
      {\label{6191612}
  \centering \includegraphics[width=0.45\linewidth]{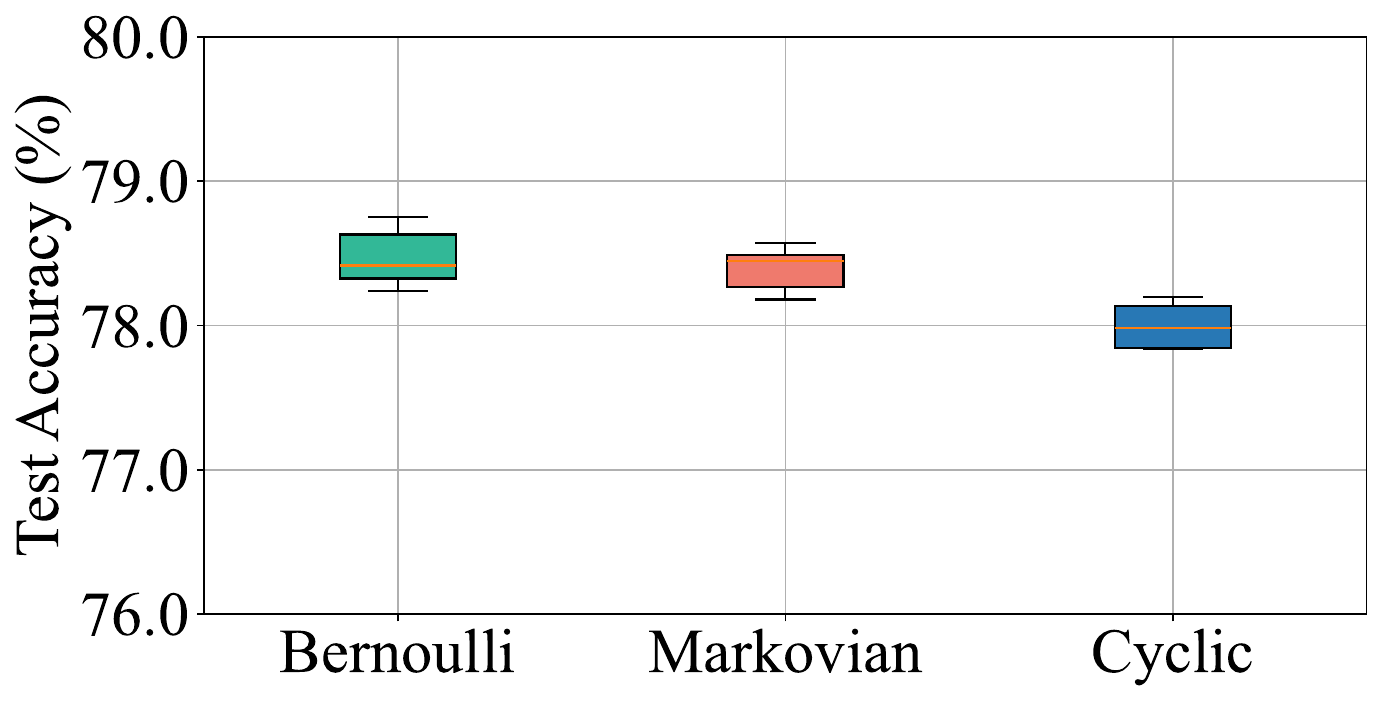}
  }
      \subfloat[CINIC]
      {\label{6191613}
  \centering \includegraphics[width=0.45\linewidth]{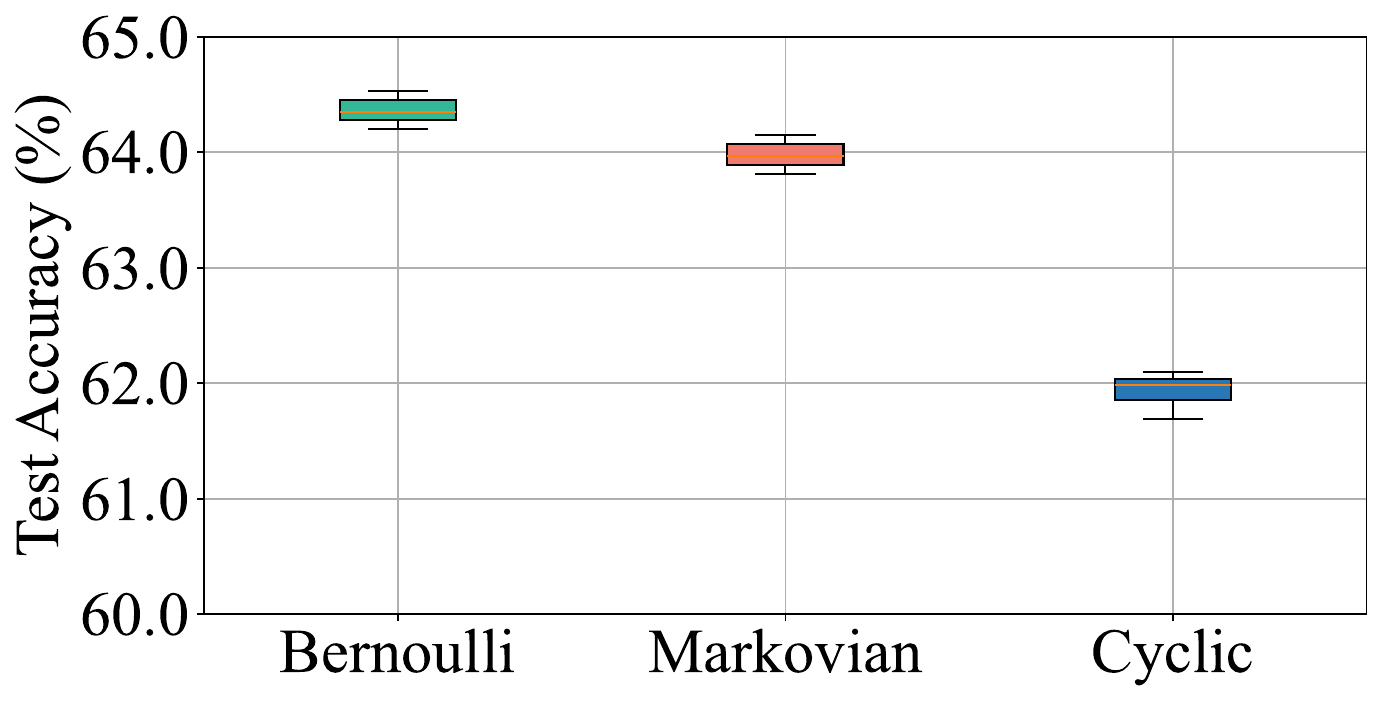}
  }
\caption{Stability of test accuracy of PMFL on CIFAR10 and CINIC under different participation patterns.}
\label{619161211}
\end{figure}

To evaluate the robustness of PMFL, it is essential to examine its stability across repeated experiments.
To this end, we conduct  experiments on  CIFAR10 and CINIC, using three participation patterns.
Specifically, for each experimental setting, we repeat  the experiment  10 times.
We record the test accuracy of the global model at the final round of each run and use these values to assess the stability of PMFL.
The test accuracy results are shown in the  Fig. \ref{619161211}.
It can be observed that  the performance variation  of PMFL on both  CIFAR10 and CINIC is within 0.5\%, indicating that  PMFL   maintains  consistent results under different participation scenarios.
Notably, no outliers are observed in these experiments, which further validates the stability of PMFL.

\subsubsection{Scalability of PMFL}
\begin{figure}[t]
\centering
    \subfloat[Bernoulli]
      {\label{53021151}
  \centering \includegraphics[width=0.45\linewidth]{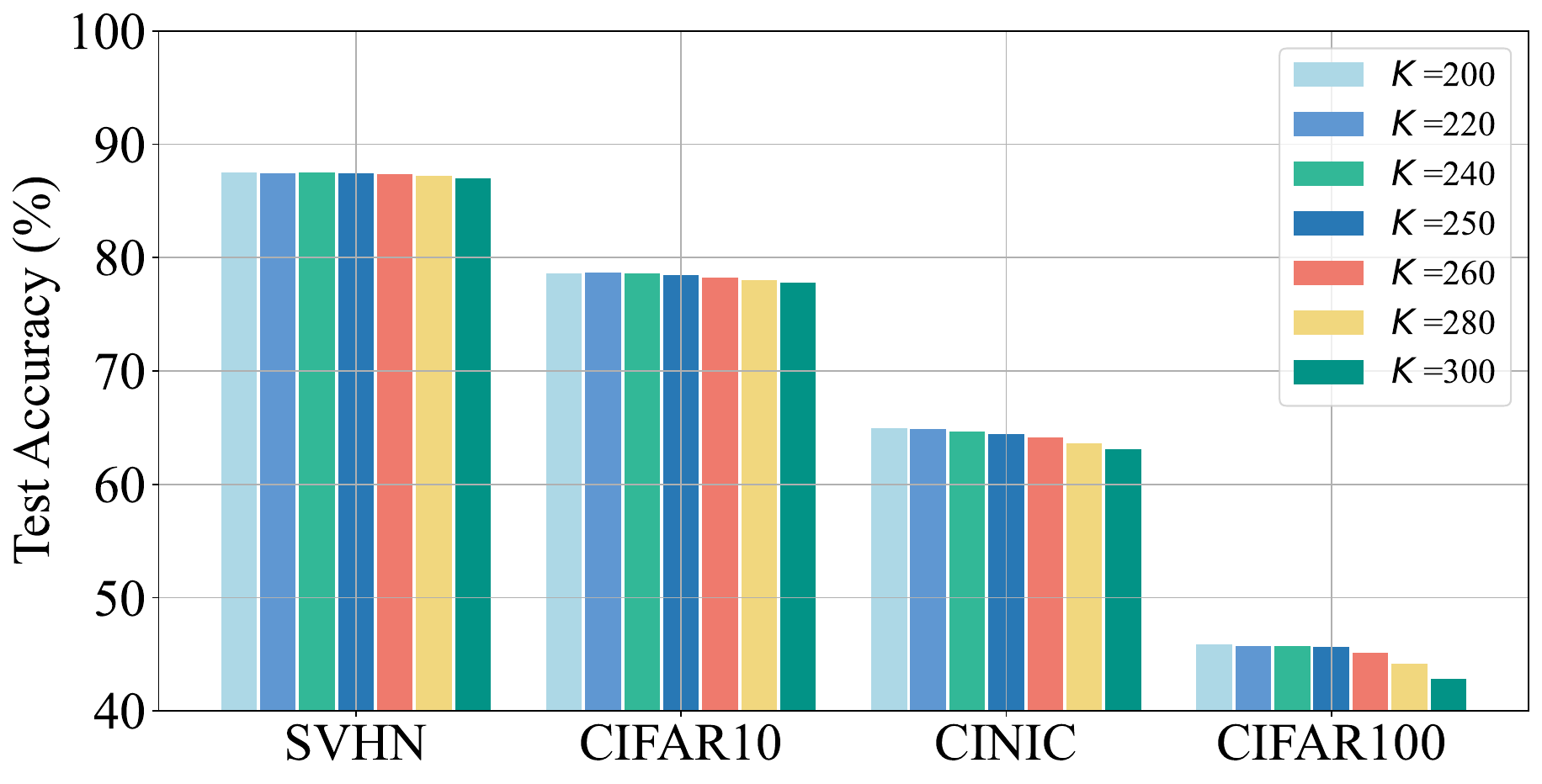}
  }
      \subfloat[Markovian]
      {\label{53021152}
  \centering \includegraphics[width=0.45\linewidth]{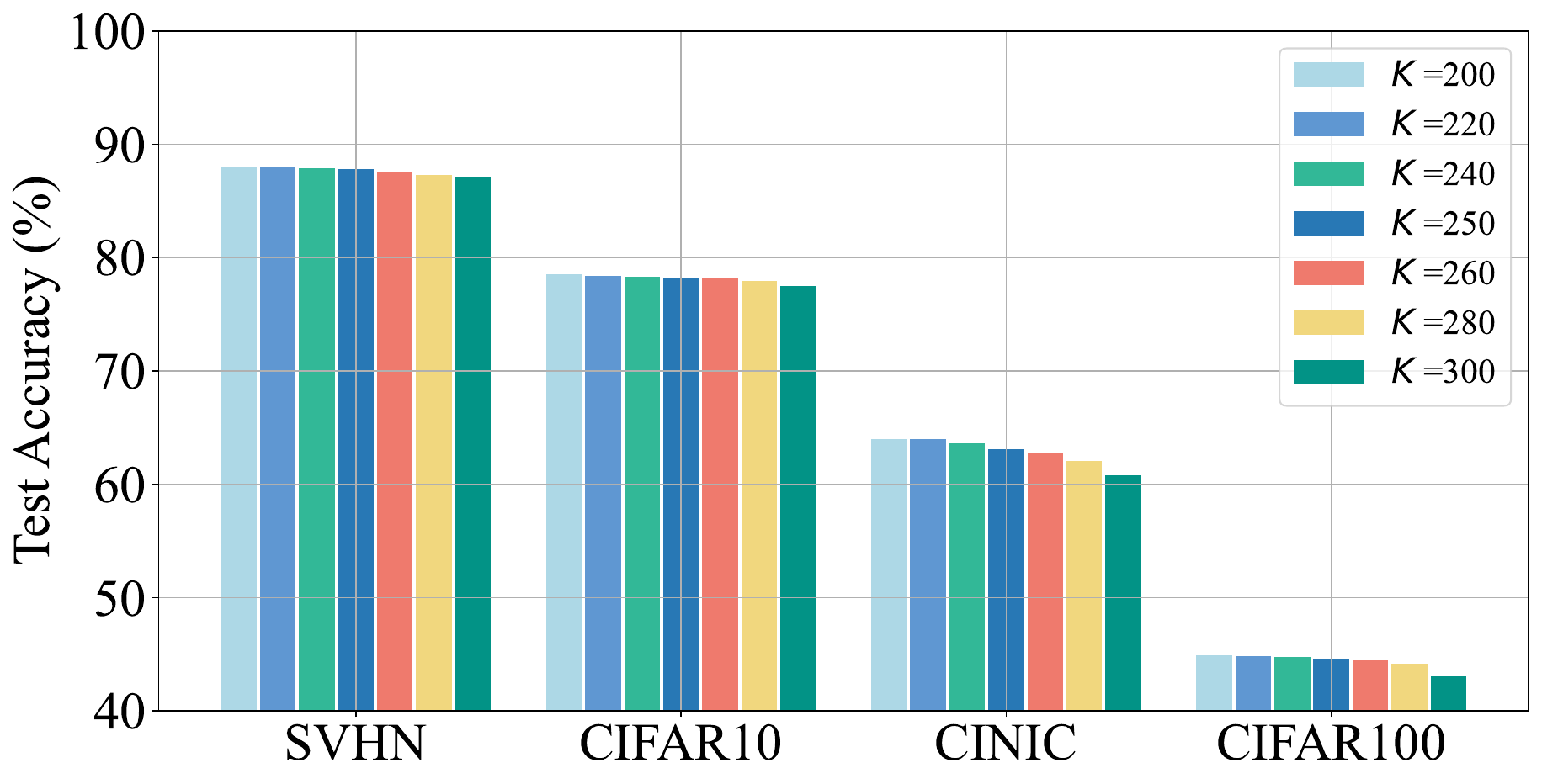}
  }
\caption{Test accuracy of PMFL on SVHN, CIFAR10, CINIC, and CIFAR100 with different numbers of nodes.}
\label{5302115}
\end{figure}

In practical FL scenarios, the total number of nodes is  uncertain.
To evaluate the scalability of PMFL, we test  its performance with  different numbers of nodes.
Specifically, we set the total number of nodes in the FL system to
$K$$\in$$\{200, 220, 240, 250, 260, 280, 300\}$.
 The experiments  are conducted under Bernoulli and Markovian patterns.
The  results are shown in Fig. \ref{5302115}.
It can be observed that the test accuracy of PMFL decreases slightly as the number of nodes gradually increases.
This phenomenon stems from the increased uncertainty in local data distributions and participation frequencies caused by the expansion of system nodes, which slightly impacts PMFL's learning performance.
Overall, PMFL maintains stable   accuracy across all  configurations, highlighting its scalability.

\subsubsection{Time Cost}

\begin{figure}[t]
  \centering
  \subfloat[Local model training time]
  {
      \label{6191713}  \includegraphics[width=0.48\linewidth]{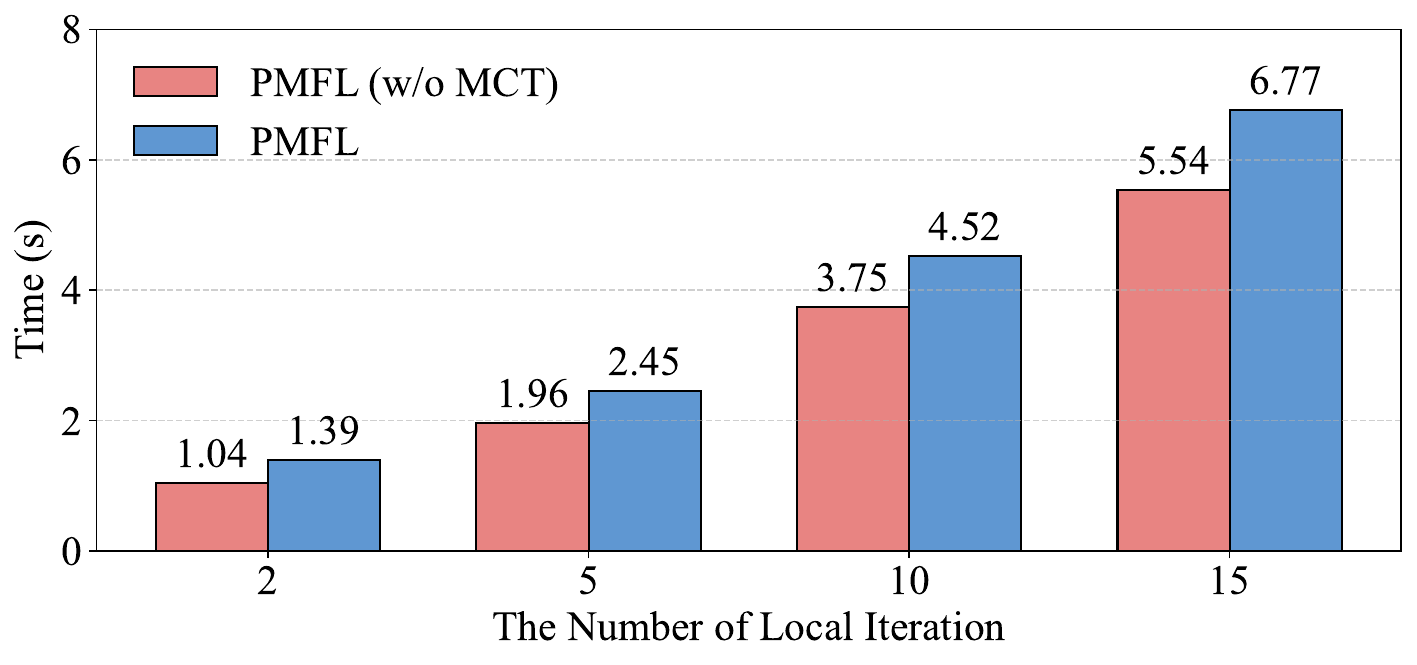}
  }
  \subfloat[Aggregation time]
  {
      \label{6191714}  \includegraphics[width=0.48\linewidth]{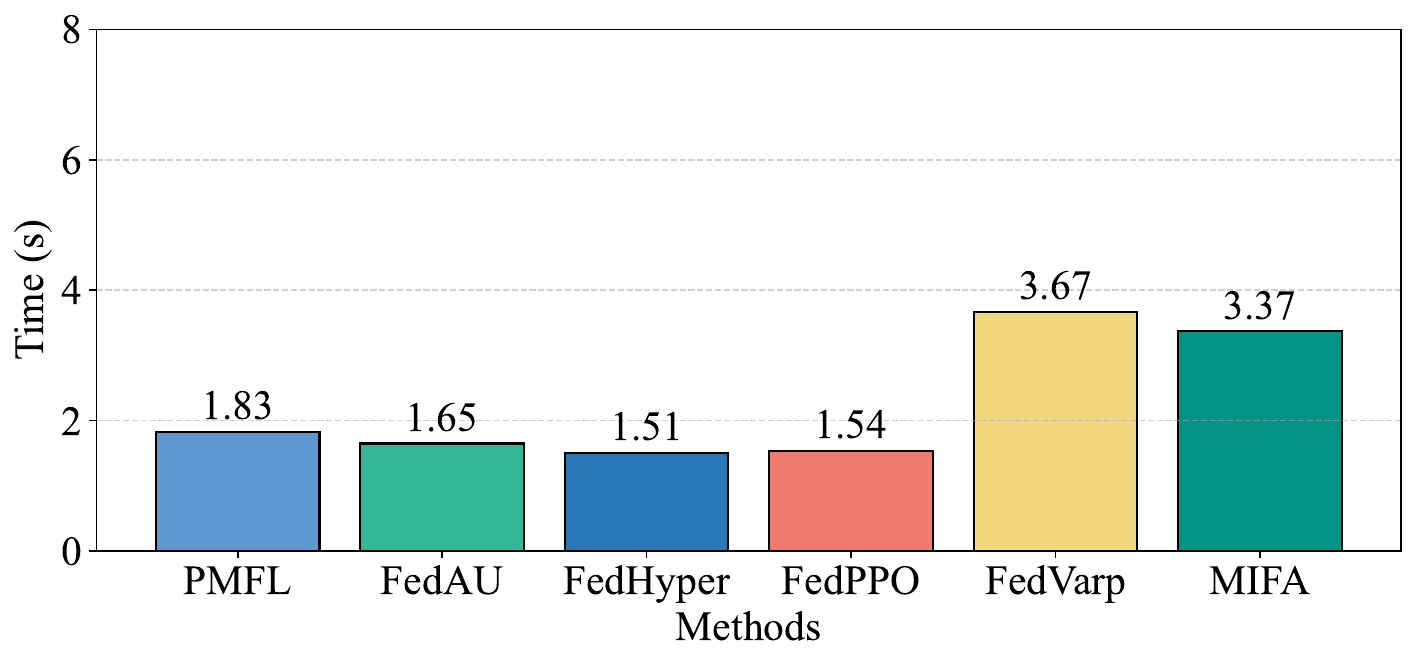}
  }
\caption{Time cost of different  methods in SVHN dataset.}
\label{61917141}
\end{figure}

To  evaluate the time cost of PMFL, we measure its  training time and aggregation time.
Specifically, we compare the per-round training time of PMFL with that of PMFL (w/o MCT), where PMFL (w/o MCT) denotes a variant of PMFL whose local function is defined without the model-contrastive term. 
All experiments are conducted on an NVIDIA A800 GPU.
The  training  times of PMFL (w/o MCT) and PMFL on the CIFAR10  are presented in Fig. \ref{6191713}.
It can be observed that, PMFL (w/o MCT) exhibits a shorter training time than PMFL,  regardless of the number of local iterations.
This is because PMFL incorporates the model-contrastive term into its optimization function, resulting in additional computational  cost.
Moreover, we  compare  the aggregation time of PMFL with that of compared methods within a single communication round, as shown in Fig. \ref{6191714}.
We notice  that FedVarp and MIFA  take the longest time to aggregate because they aggregate model updates from all nodes, including both  participating nodes and non-participating nodes.
Moreover, we observe  that FedHyper and FedPPO achieve  the shortest aggregation time since they operate only on the model updates of currently participating nodes.
Compared with these methods, PMFL take  a slightly higher aggregation cost   because the server  computes  aggregation weights for each node and incorporates historical global models. 
Since PMFL improves the global model performance under heterogeneous scenarios, this additional time cost remains within an acceptable range.

\section{CONCLUSION} \label{62314441}
In this paper, we propose PMFL,  a performance-enhanced model-contrastive federated learning framework that improves the FL performance in heterogeneous scenarios.
PMFL leverages historical local models into the model-contrastive term of the optimization function to construct stable contrastive points, thereby enhancing the
effectiveness of model-contrastive term.
Moreover, PMFL  employs the aggregation strategy that adaptively adjusts aggregation weights and incorporates historical global models.
Extensive experiments validate the effectiveness and superiority of  PMFL compared to other  methods under heterogeneous scenarios.
For future work, we plan to investigate model heterogeneity in FL.

\bibliographystyle{IEEEtran}
\bibliography{ref}

@article{miao2023secure,
  title={Secure model-contrastive federated learning with improved compressive sensing},
  author={Miao, Yinbin and Zheng, Wei and Li, Xinghua and Li, Hongwei and Choo, KimKwang Raymond and Deng, Robert H},
  journal={IEEE Trans. Inf. Forensics Security}, 
  volume={18},
  pages={3430--3444},
  year={2023},
}

@article{bonawitz2019towards,
  title={Towards federated learning at scale: System design},
  author={Bonawitz, Keith and  others},
  journal={Proc.  Mach. Learn. Syst.},
  volume={1},
  pages={374--388},
  year={2019}
}

@inproceedings{jhunjhunwala2022fedvarp,
  title={Fedvarp: Tackling the variance due to partial client participation in federated learning},
  author={Jhunjhunwala, Divyansh and Sharma, Pranay and Nagarkatti, Aushim and Joshi, Gauri},
  booktitle={Proc. Uncertainty Artif. Intell.},
  pages={906--916},
  year={2022},
}

@inproceedings{gu2021fast,
  title={Fast federated learning in the presence of arbitrary device unavailability},
  author={Gu, Xinran and Huang, Kaixuan and Zhang, Jingzhao and Huang, Longbo},
 booktitle = {Proc. Adv. Neural Inf. Process. Syst.},
  volume={34},
  pages={12052--12064},
  year={2021}
}

@inproceedings{wang2023lightweight,
  title={A Lightweight Method for Tackling Unknown Participation Statistics in Federated Averaging},
  author={Wang, Shiqiang and Ji, Mingyue},
  booktitle={Proc. Int. Conf. Learn. Representations},
  year={2024}
}

@inproceedings{li2021model,
  title={Model-contrastive federated learning},
  author={Li, Qinbin and He, Bingsheng and Song, Dawn},
  booktitle={Proc. IEEE/CVF Conf. Comput. Vis. Pattern Recognit.},
  pages={10713--10722},
  year={2021}
}

@InProceedings{Seo_2024_CVPR,
    author    = {Seo, Seonguk and Kim, Jinkyu and Kim, Geeho and Han, Bohyung},
    title     = {Relaxed Contrastive Learning for Federated Learning},
  booktitle={Proc. IEEE/CVF Conf. Comput. Vis. Pattern Recognit.},
    year      = {2024},
    pages     = {12279-12288}
}

@article{zhang2025swim,
  title={SWIM: Sliding-Window Model contrast for federated learning},
  author={Zhang, Heng-Ru and Chen, Rui and Wen, Shi-Huai and Bian, Xiao-Qiang},
  journal={Future Gener. Comput. Syst.},
  volume={164},
  number={107590},
  year={2025},
}

@ARTICLE{9847055,
  author={Deng, Yongheng and others},
  journal={IEEE Trans. Parallel Distrib. Syst.}, 
  title={Improving Federated Learning With Quality-Aware User Incentive and Auto-Weighted Model Aggregation}, 
  year={2022},
  volume={33},
  number={12},
  pages={4515-4529},
}

@InProceedings{pmlr-v202-li23s,
  title = 	 {Revisiting Weighted Aggregation in Federated Learning with Neural Networks},
  author =       {Li, Zexi and Lin, Tao and Shang, Xinyi and Wu, Chao},
  booktitle={Proc. Int. Conf. Mach. Learn.},
  pages = 	 {19767--19788},
  year = 	 {2023},
 volume = 	 {202},
}

@ARTICLE{9425020,
  author={Chen, Shengbo and Shen, Cong and Zhang, Lanxue and Tang, Yuanmin},
  journal={IEEE Trans. Wireless Commun.}, 
  title={Dynamic Aggregation for Heterogeneous Quantization in Federated Learning}, 
  year={2021},
  volume={20},
  number={10},
  pages={6804-6819},
}

@ARTICLE{10854512,
  author={Li, Xiaodong and Gao, Yulong and Deng, Yansha and Jiang, Xinzhuo},
  journal={IEEE Trans. Cognit. Commun. Networking}, 
  title={Federated Learning With Adaptive Aggregation Weights for Non-IID Data in Edge Networks}, 
  year={2025},
  volume={},
  number={},
  pages={1-1},
}

@inproceedings{zheng2024heterogeneous,
  title={Heterogeneous contrastive learning for foundation models and beyond},
  author={Zheng, Lecheng and Jing, Baoyu and Li, Zihao and Tong, Hanghang and He, Jingrui},
  booktitle={Proc.  ACM  Conf. Knowl. Discov. Data Mining},
  pages={6666--6676},
  year={2024}
}

@article{liu2021self,
  title={Self-supervised learning: Generative or contrastive},
  author={Liu, Xiao and others},
  journal={IEEE Trans. Knowl. Data Eng.}, 
  volume={35},
  number={1},
  pages={857--876},
  year={2021},
}

@inproceedings{netzer2011reading,
  title={Reading digits in natural images with unsupervised feature learning},
  author={Netzer, Yuval  and others},
  booktitle={Proc. Int. Conf. Neural Inf. Process. Syst. Workshop},
  year={2011},
}

@article{krizhevsky2009learning,
  title={Learning multiple layers of features from tiny images},
  author={Krizhevsky, Alex and Hinton, Geoffrey and others},
  year={2009},
  publisher={Toronto, ON, Canada},
 url = {http://www.cs.utoronto.ca/~kriz/learning-features-2009-TR.pdf}
}

@article{darlow2018cinic,
  title={Cinic-10 is not imagenet or cifar-10},
  author={Darlow, Luke N and Crowley, Elliot J and Antoniou, Antreas and Storkey, Amos J},
   journal= {arXiv preprint arXiv:1810.03505},
  year={2018}
}

@ARTICLE{mu2024feddmc,
  author={Mu, Xutong and others},
  journal={IEEE Trans. Dependable Secure Comput.}, 
  title={FedDMC: Efficient and Robust Federated Learning via Detecting Malicious Clients}, 
  year={2024},
  volume={21},
  number={6},
  pages={5259-5274},
}

@ARTICLE{10549523,
  author={Zhang, Xinyu and others},
  journal={IEEE Trans. Inf. Forensics Security}, 
  title={FLTracer: Accurate Poisoning Attack Provenance in Federated Learning}, 
  year={2024},
  volume={19},
  number={},
  pages={9534-9549},
}

@inproceedings{mcmahan2017communication,
  title={Communication-efficient learning of deep networks from decentralized data},
  author={McMahan, Brendan and Moore, Eider and Ramage, Daniel and Hampson, Seth and y Arcas, Blaise Aguera},
  booktitle={Proc. Artif. Intell. Statist.},
  pages={1273--1282},
  year={2017},
}

@ARTICLE{10468591,
  author={Lu, Zili and Pan, Heng and Dai, Yueyue and Si, Xueming and Zhang, Yan},
  journal={IEEE Internet Things J.}, 
  title={Federated Learning With Non-IID Data: A Survey}, 
  year={2024},
  volume={11},
  number={11},
  pages={19188-19209},
}

@inproceedings{khosla2020supervised,
  title={Supervised contrastive learning},
  author={Khosla, Prannay and others},
  booktitle={Proc. Adv. Neural Inf. Process. Syst.},
  volume={33},
  pages={18661--18673},
  year={2020}
}

@ARTICLE{10108910,
  author={Albaseer, Abdullatif Mohammed and Abdallah, Mohamed and Al-Fuqaha, Ala and Seid, Abegaz Mohammed and Erbad, Aiman and Dobre, Octavia A.},
  journal={IEEE Trans. Netw. Serv. Manage.}, 
  title={Fair Selection of Edge Nodes to Participate in Clustered Federated Multitask Learning}, 
  year={2023},
  volume={20},
  number={2},
  pages={1502-1516},
}

@ARTICLE{10001832,
  author={Yang, Kun and Chen, Shengbo and Shen, Cong},
  journal={IEEE J. Sel. Areas Commun.}, 
  title={On the Convergence of Hybrid Server-Clients Collaborative Training}, 
  year={2023},
  volume={41},
  number={3},
  pages={802-819},
}

@ARTICLE{10138783,
  author={Albaseer, Abdullatif and Abdallah, Mohamed and Al-Fuqaha, Ala and Erbad, Aiman},
  journal={IEEE Trans. Syst., Man, Cybern. Syst.}, 
  title={Data-Driven Participant Selection and Bandwidth Allocation for Heterogeneous Federated Edge Learning}, 
  year={2023},
  volume={53},
  number={9},
  pages={5848-5860},
}

@ARTICLE{10528890,
  author={Luo, Long and Zhang, Chi and Yu, Hongfang and Sun, Gang and Luo, Shouxi and Dustdar, Schahram},
  journal={IEEE Trans. Serv. Comput.}, 
  title={Communication-Efficient Federated Learning With Adaptive Aggregation for Heterogeneous Client-Edge-Cloud Network}, 
  year={2024},
  volume={17},
  number={6},
  pages={3241-3255},
}

@inproceedings{NEURIPS2020564127c0,
 author = {Wang, Jianyu and Liu, Qinghua and Liang, Hao and Joshi, Gauri and Poor, H. Vincent},
 booktitle = {Proc. Adv. Neural Inf. Process. Syst.},
 pages = {7611--7623},
 title = {Tackling the Objective Inconsistency Problem in Heterogeneous Federated Optimization},
 volume = {33},
 year = {2020}
}

@misc{ying2025exactlinearconvergencefederated,
      title={Exact and Linear Convergence for Federated Learning under Arbitrary Client Participation is Attainable}, 
      author={Bicheng Ying and Zhe Li and Haibo Yang},
      year={2025},
      eprint={2503.20117},
      archivePrefix={arXiv},
      primaryClass={cs.LG},
      url={https://arxiv.org/abs/2503.20117}, 
}

@inproceedings{NEURIPS2024_bcaebb60,
 author = {Xiang, Ming and Ioannidis, Stratis and Yeh, Edmund and Joe-Wong, Carlee and Su, Lili},
booktitle = {Proc. Adv. Neural Inf. Process. Syst.},
 pages = {104281--104328},
 title = {Efficient Federated Learning against Heterogeneous and Non-stationary Client Unavailability},
 volume = {37},
 year = {2024}
}

@misc{weng2025heterogeneityawareclientsamplingunified,
      title={Heterogeneity-Aware Client Sampling: A Unified Solution for Consistent Federated Learning}, 
      author={Shudi Weng and Chao Ren and Ming Xiao and Mikael Skoglund},
      year={2025},
      eprint={2505.11304},
      archivePrefix={arXiv},
      primaryClass={cs.LG},
      url={https://arxiv.org/abs/2505.11304}, 
}

@ARTICLE{10413546,
  author={Yang, Yuning and Liu, Xiaohong and Gao, Tianrun and Xu, Xiaodong and Zhang, Ping and Wang, Guangyu},
  journal={IEEE Journal of Biomedical and Health Informatics}, 
  title={Dense Contrastive-Based Federated Learning for Dense Prediction Tasks on Medical Images}, 
  year={2024},
  volume={28},
  number={4},
  pages={2055-2066},
}

@ARTICLE{10286887,
  author={Zhou, Tailin and Zhang, Jun and Tsang, Danny H. K.},
 journal={IEEE Trans. Mobile Comput.},
  title={FedFA: Federated Learning With Feature Anchors to Align Features and Classifiers for Heterogeneous Data}, 
  year={2024},
  volume={23},
  number={6},
  pages={6731-6742},
}

@ARTICLE{11075614,
  author={Xu, Bin and others},
 journal={IEEE Trans. Mobile Comput.},
  title={Heterogeneous Federated Learning driven by Multi-Knowledge Distillation}, 
  year={2025},
  volume={},
  number={},
  pages={1-14},
  doi={10.1109/TMC.2025.3586921}}

@ARTICLE{10556806,
  author={Ning, Wanyi and others},
journal={IEEE Trans. Serv. Comput.},
  title={One Teacher is Enough: A Server-Clueless Federated Learning With Knowledge Distillation}, 
  year={2024},
  volume={17},
  number={5},
  pages={2704-2718},
}

@ARTICLE{10476711,
  author={Liao, Guocheng and Luo, Bing and Feng, Yutong and Zhang, Meng and Chen, Xu},
 journal={IEEE Trans. Mobile Comput.},
  title={Optimal Mechanism Design for Heterogeneous Client Sampling in Federated Learning}, 
  year={2024},
  volume={23},
  number={11},
  pages={10598-10609},
}

@ARTICLE{10443546,
  author={Luo, Bing and Xiao, Wenli and Wang, Shiqiang and Huang, Jianwei and Tassiulas, Leandros},
journal={IEEE Trans. Mobile Comput.},
  title={Adaptive Heterogeneous Client Sampling for Federated Learning Over Wireless Networks}, 
  year={2024},
  volume={23},
  number={10},
  pages={9663-9677},
}

@inproceedings{xu2025federated,
  title={Federated learning with sample-level client drift mitigation},
  author={Xu, Haoran and Li, Jiaze and Wu, Wanyi and Ren, Hao},
booktitle={Proc. AAAI Conf. Artif. Intell.},
  volume={39},
  number={20},
  pages={21752--21760},
  year={2025}
}

@article{lin2016dirichlet,
  title={On the dirichlet distribution},
  author={Lin, Jiayu},
  journal={Master’s thesis, Dept. Math. Statist., Queen's Univ.},
  volume={40},
  year={2016}
}

@article{gu2018recent,
  title={Recent advances in convolutional neural networks},
  author={Gu, Jiuxiang and  others},
  journal={Pattern Recognit.},
  volume={77},
  pages={354--377},
  year={2018},
}

@article{zhang2017residual,
  title={Residual networks of residual networks: Multilevel residual networks},
  author={Zhang, Ke and Sun, Miao and Han, Tony X and Yuan, Xingfang and Guo, Liru and Liu, Tao},
  journal={IEEE Trans. Circuits Syst. Video Technol.},
  volume={28},
  number={6},
  pages={1303--1314},
  year={2017},
  publisher={IEEE}
}

@ARTICLE{11202428,
  author={Zhang, Hongliang and others},
  journal={IEEE Trans. Dependable Secure Comput.}, 
  title={Towards Model-Contrastive Federated Learning with Lightweight Privacy Preservation and Poisoning Attack Detection}, 
  year={2025},
  pages={1-17},
  doi={10.1109/TDSC.2025.3620529}}

@inproceedings{wang2023fedhyper,
  title={Fedhyper: A universal and robust learning rate scheduler for federated learning with hypergradient descent},
  author={Wang, Ziyao and Wang, Jianyu and Li, Ang},
  booktitle={Proc. Int. Conf. Learn. Representations},
  year={2024}
}

@ARTICLE{10909702,
  author={Zhao, Zheyu and Li, Anran and Li, Ruidong and Yang, Lei and Xu, Xiaohua},
  journal={IEEE Trans. Cognit. Commun. Networking.}, 
  title={FedPPO: Reinforcement Learning-Based Client Selection for Federated Learning With Heterogeneous Data}, 
  year={2025},
  volume={11},
  number={6},
  pages={4141-4153},
  doi={10.1109/TCCN.2025.3547751}}

@ARTICLE{11192608,
  author={Zhang, Xuning and Li, Jian and Yin, Rong and Wang, Weiping},
  journal={IEEE Trans. Neural Netw. Learn. Syst.}, 
  title={FedNK-RF: Federated Kernel Learning With Heterogeneous Data and Optimal Rates}, 
  year={2025},
  volume={},
  number={},
  pages={1-13},
  doi={10.1109/TNNLS.2025.3612728}}

@ARTICLE{10197242,
  author={Huang, Honglan and Shi, Wei and Feng, Yanghe and Niu, Chaoyue and Cheng, Guangquan and Huang, Jincai and Liu, Zhong},
  journal={IEEE Trans. Neural Netw. Learn. Syst.}, 
  title={Active Client Selection for Clustered Federated Learning}, 
  year={2024},
  volume={35},
  number={11},
  pages={16424-16438},
  doi={10.1109/TNNLS.2023.3294295}}

@ARTICLE{10816699,
  author={Wu, Zheshun and Xu, Zenglin and Zeng, Dun and Wang, Qifan and Liu, Jie},
  journal={IEEE Trans. Neural Netw. Learn. Syst.}, 
  title={Advocating for the Silent: Enhancing Federated Generalization for Nonparticipating Clients}, 
  year={2025},
  volume={36},
  number={8},
  pages={14174-14188},
  doi={10.1109/TNNLS.2024.3516692}}
\end{document}